\documentclass[lettersize,journal]{IEEEtran}

\IEEEoverridecommandlockouts                              % This command is only needed if 
\usepackage{lipsum}
\usepackage{colortbl}
\usepackage{url}
\usepackage{graphicx}
\usepackage{amsmath}
\usepackage{amssymb}
\usepackage{multirow}
\usepackage{array}
\usepackage[linesnumbered,ruled,vlined]{algorithm2e}
\usepackage{booktabs}
\usepackage[noadjust]{cite}
\usepackage{soul}
\usepackage[colorlinks,allcolors=black]{hyperref}
\renewcommand\hl[1]{#1}
\ifCLASSINFOpdf
\else
\fi
\ifCLASSOPTIONcompsoc
\else
  \usepackage[caption=false,font=footnotesize]{subfig}
\fi
\begin{document}

\title{IPAPRec: A promising tool for learning high-performance mapless navigation skills with deep reinforcement learning}

% \author{Michael~Shell,~\IEEEmembership{Member,~IEEE,}
%         John~Doe,~\IEEEmembership{Fellow,~OSA,}
%         and~Jane~Doe,~\IEEEmembership{Life~Fellow,~IEEE}% <-this % stops a space
% \thanks{M. Shell was with the Department
% of Electrical and Computer Engineering, Georgia Institute of Technology, Atlanta,
% GA, 30332 USA e-mail: (see http://www.michaelshell.org/contact.html).}% <-this % stops a space
% \thanks{J. Doe and J. Doe are with Anonymous University.}% <-this % stops a space
% \thanks{Manuscript received April 19, 2005; revised August 26, 2015.}}
\author{Wei Zhang, Yunfeng Zhang, Ning Liu, Kai Ren and Pengfei Wang
\thanks{ This article has been accepted for publication in IEEE/ASME Transactions on Mechatronics. This is the author's version which has not been fully edited and
content may change prior to final publication. Citation information: DOI 10.1109/TMECH.2022.3182427 \textit{(Corresponding author: Kai Ren.)}}%
\thanks{\textcopyright 2022 IEEE. Personal use of this material is permitted.  Permission from IEEE must be obtained for all other uses, in any current or future media, including reprinting/republishing this material for advertising or promotional purposes, creating new collective works, for resale or redistribution to servers or lists, or reuse of any copyrighted component of this work in other works.}%
\thanks{Wei Zhang and Yunfeng Zhang are with the Department of Mechanical Engineering, National University of Singapore.
	{\tt\small e-mail: weizhang@u.nus.edu, mpezyf@nus.edu.sg}
	}%
\thanks{Ning Liu are with the Smart Manufacturing Division, Advanced Remanufacturing and Technology Centre, Singapore.
	{\tt\small e-mail:Liu\_Ning@artc.a-star.edu.sg}
	}%
\thanks{Kai Ren are with the School of Mechanical Engineering, Zhejiang University, China.
	{\tt\small e-mail:renkai@u.nus.edu}
	}%
\thanks{Pengfei Wang are with the Temasek Laboratories, National University of Singapore.
	{\tt\small e-mail:wangpengfei@u.nus.edu}
	}%
}
\markboth{IEEE/ASME TRANSACTIONS ON MECHATRONICS.}
{Wei Zhang \MakeLowercase{\textit{et al.}}: IPAPRec: A promising tool for learning high-performance mapless navigation skills with deep reinforcement learning}
\maketitle
\begin{abstract}
This paper studies how to improve the generalization performance and learning speed of the navigation agents trained with deep reinforcement learning (DRL). Although DRL exhibits huge potential in robot mapless navigation, DRL agents performing well in training scenarios are often found to perform poorly in unfamiliar scenarios. In this work, we propose that the representation of LiDAR readings is a key factor behind the degradation of agents' performance and present a powerful input pre-processing (IP) approach to address this issue. As this approach uses adaptively parametric reciprocal functions to pre-process LiDAR readings, we refer to this approach as IPAPRec and its normalized version as IPAPRecN. IPAPRec/IPAPRecN can highlight important short-distance values and compress the range of less-important long-distance values in laser scans, which well address the issues induced by conventional representations of laser scans. Their high performance \hl{was} validated by extensive simulation and real-world experiments. The results show that our methods can substantially improve navigation agents’ generalization performance and greatly reduce the training time compared to conventional methods.
\end{abstract}

% \begin{IEEEkeywords}
% IEEE, IEEEtran, journal, \LaTeX, paper, template.
% \end{IEEEkeywords}

% \fi
\IEEEpeerreviewmaketitle

\section{Introduction}
% for IEEE journal papers produced under \LaTeX\ using
% IEEEtran.cls version 1.8b and later.
\IEEEPARstart{A}{utonomous} navigation is a basic requirement for mobile robots. Compared with map-based approaches \cite{Yue2021}, mapless navigation is useful for performing tasks in unknown and dynamic scenarios, such as search and rescue, where prior maps are not available and the layout of the scenarios changes continually \cite{Pfeiffer2018}. Recently, deep reinforcement learning (DRL) \cite{Mnih2015} has emerged as a promising technique for learning mapless navigation skills. In DRL, the agent for the mobile robot utilizes deep neural networks (DNN) \cite{Lecun2015} to learn navigation skills within the framework of reinforcement learning. Trained with DRL, the navigation agent was found to outperform their conventional counterparts, such as dynamic window approach (DWA) \cite{Fox1997} and artificial potential fields (APF) \cite{warren1989global}, in local path planning \cite{Tai2017,rana2020residual}. It \hl{could} quickly adapt to the dynamic obstacles during navigation \cite{Pfeiffer2018,zhang2021learn,Everett2021}, which is \hl{essentially required} for navigation in pedestrian-rich scenarios.

Learning navigation skills with DRL typically requires a considerable number of training episodes, and each episode contains hundreds of robot-environment interaction steps \cite{Wu2019}. To avoid \hl{damages of the robot occurred} in real-world collisions and accelerate training, it is a common practice to train the DRL agent in simulation and then deploy it to a real robot \cite{Hu2021,Wu2021}. In simulation, 2D LiDAR sensors are commonly used for obstacle detection \cite{Tai2017,zhai2021}, because they can be effectively simulated in lightweight simulators, such as \textit{ROS Stage} \cite{Stage_rosWiki} and \textit{Box2D} \cite{Box2D:Games}, with high fidelity.  Moreover, distance values captured by 3D sensors (e.g., depth cameras) \hl{can} be easily integrated into the inputs of 2D LiDAR, thus allowing robots trained with 2D LiDAR to avoid obstacles of different heights \cite{Leiva2020}.

\begin{figure}[t]
    \centering
	  \subfloat[\cite{Tai2017}]{
       \includegraphics[width=0.4\linewidth]{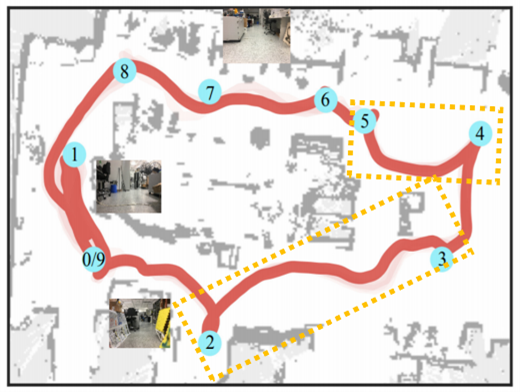}\label{e1}}
	  \subfloat[\cite{luong2021incremental}]{
        \includegraphics[width=0.4\linewidth]{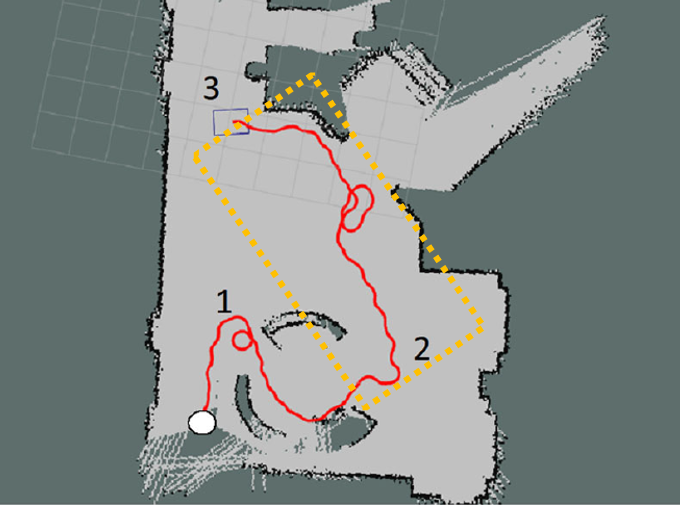}\label{e2}}
        \\[-2ex]
	    \subfloat[\cite{Leiva2020}]{
        \includegraphics[width=0.4\linewidth]{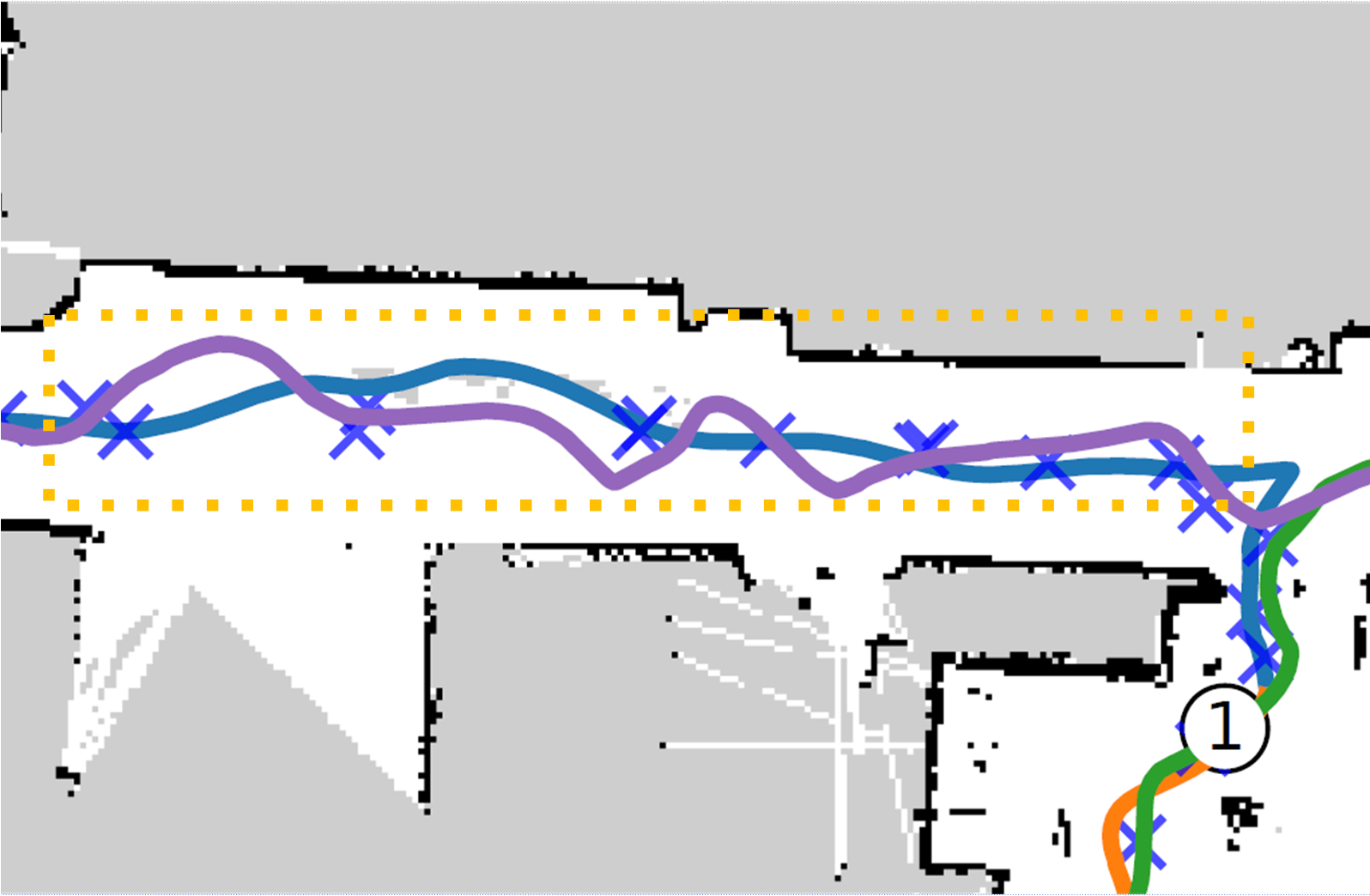}\label{e4}}
	  \subfloat[\cite{Dobrevski2020}]{
        \includegraphics[width=0.4\linewidth]{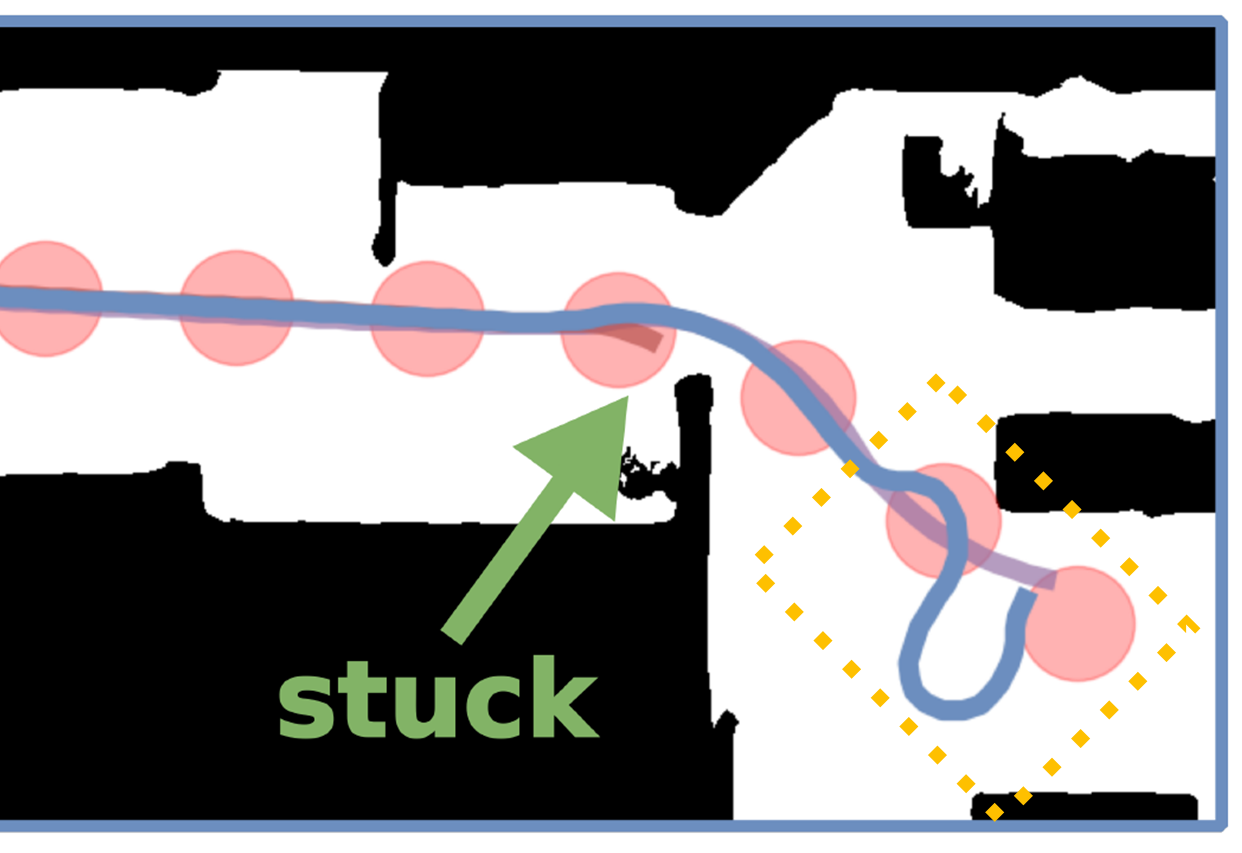}\label{e3}}
	\caption{Examples of abnormal behaviors of DRL navigation agent when tested in real-world unseen environments.}
	\label{examples}
\end{figure}

Currently, most of the studies in this field \hl{focused} on improving the agent's navigation performance in training scenarios \cite{Xie2018,xie2021}. Accordingly, good navigation results can be achieved when the testing scenarios \hl{were} similar to the training scenarios \cite{Shi2020}. However, when tested in unfamiliar scenarios, some abnormal behaviors of the DRL agent \hl{were} often observed \cite{Tai2017,luong2021incremental,Leiva2020,Dobrevski2020,Fan2020DistributedScenarios}. For example, as shown in Fig. \ref{examples}, when there are no obstacles between robots and targets (marked with orange boxes), the agents choose much longer paths instead of straight lines. Moreover, as shown in Fig. \ref{e1} and \ref{e2}, the robots even tend to approach obstacles during the goal-reaching process. Fan et al. \cite{Fan2020DistributedScenarios} also observed their robot \hl{wandered} around a nearby target instead of approaching it.

The above evidences reveal that the DRL navigation agents may behave anomalously and are not reliable in real-world scenarios. Fan et al. \cite{Fan2020DistributedScenarios} attributed the anomalous behaviors of the DRL agent to the learning scheme, i.e., the DRL agent \hl{failed} to collect training data in some special situations and \hl{had} few opportunities to learn optimal actions; the exploration noise used in training made it challenging for the agent to learn deterministic optimal policies. However, they ignored the uniqueness of DRL-based robot navigation, i.e., the input representations of LiDAR readings.

In most of the previous works, the LiDAR readings \hl{were} represented \hl{as} a vector, where each element \hl{could} be either raw distance value \cite{Shi2020,Lim2020} or linearly normalized distance value \cite{Tai2017,Pfeiffer2018} in a specific direction. In this paper, we refer to these input representations of LiDAR readings as conventional representations. \hl{Besides, few works {\cite{pfeiffer2018data}}, {\cite{liu2020robot}} utilized the 2D occupancy grid as the input representation of the LiDAR 
readings. However, to reduce the dimension of the input, the resolution of 2D occupancy grid is much lower (0.1m per cell in {\cite{pfeiffer2018data}}) than that of the 1D representation, which is not suitable in crowded scenarios. In addition, due to the large input space of the 2D occupancy grid, it requires additional pre-training of a convolutional auto-encoder to extract useful features before training the DRL agent. Although the conventional representations have relatively high resolution and  small input space,} they pose two challenges to the learning agent. \hl{Firstly}, such representations require the policy network to generate two completely different policies on similar inputs. \hl{Secondly}, the major part of input space is occupied by long-distance values. Such values have little impact on obstacle avoidance, but it is necessary for the agent to collect enough training samples containing such values to enhance its performance in relatively open scenarios. To address \hl{these} two issues, in this work, a novel input representation of LiDAR readings is proposed. The new representation can be obtained by using \hl{an} input pre-processing (IP) approach named IPAPRec, which utilizes the adaptively parametric reciprocal function to process LiDAR input. With the new input representations, the DRL agents can learn much faster during training and can generalize better in unseen scenarios than their conventional counterparts. To sum up, the contributions of this work are as follows\hl{:}
 
\begin{itemize}
	\item \hl{The limitations of the conventional input representations of LiDAR readings for DRL-based navigation have been proposed.}
	\item A novel IP approach named IPAPRec \hl{has been} proposed to accelerate training and enhance the generalization capability of the DRL agent.
	\item Extensive experiments have been carried out and the corresponding results \hl{have verified} the high efficiency of our IPAPRec approach.
\end{itemize}

The remainder of this paper is organized as follows. The related works and preliminaries are introduced in Section \ref{RW} and \ref{lim}, respectively. Afterwards, we introduce our IPAPRec approach in Section \ref{approach}, followed by model training and evaluation in Section \ref{sim}. The results of real-world experiments are presented in Section \ref{real-experiment}. Last, conclusions are given in Section \ref{conclusion}.

\section{Related works}\label{RW}
Recently, DRL has been widely employed to solve mapless navigation problems for mobile robots. As the interaction between the robot and environment takes up most of the training time, improving the learning speed of the DRL agent has received the most attention in prior works. The most straightforward idea to speed up training is to improve the DRL algorithm, such as changing the DDPG (deep deterministic policy gradient \cite{Lillicrap2016}) algorithm to a more powerful algorithm named SAC (soft actor critic \cite{Haarnoja2018}) \cite{DeJesus2021}, adopting intrinsic reward to encourage the robot to explore more unvisited regions \cite{Shi2020}, or using asynchronous training \cite{Tai2017}. These approaches are highly dependent on the development of DRL algorithms and are not specific to robot navigation. 

To accelerate training in the context of solving robot navigation problems, transferring the expert skills to the DRL agent is a feasible option \cite{Pfeiffer2018,Lim2020}. For example, \hl{Pfeiffer et al. {\cite{Pfeiffer2018}} used millions of expert data to pre-train the DRL agent in a supervised manner.} With pre-trained skills, the agent \hl{could} learn faster than the agent purely trained with DRL. However, collecting good demonstrations requires \hl{extensive} expert experience, \hl{thus} the pre-trained agent is limited to expert performance. Instead of collecting expert data, Xie et al. \cite{xie2021} verified that using some simple navigation controllers, such as a PID controller and an obstacle-avoidance controller, \hl{could} facilitate the learning process through switching between the candidate conventional controllers and the DDPG controller.

When learning navigation skills with DRL, as previously introduced in Fig. \ref{examples}, quick learners are not guaranteed to have good generalization performance in real-world scenarios. So far, only a few works have noticed and aimed to address this issue, and most of them utilize external controllers to assist the DRL agent's decision-making. To achieve this, a high-level decision-maker needs to assess the current situation and activate the external controllers in cases \hl{when} the DRL agents fail. For example, to avoid agents' abnormal behaviors in open scenarios, the PID controller \hl{would} replace the DRL controller \cite{Fan2020DistributedScenarios} when current situation \hl{was} considered safe (no obstacles nearby). However, \hl{choosing} a feasible threshold for switching the controllers is nontrivial, which requires fine-tuning based on the specific behavior of the DRL agent. Besides, by evaluating the uncertainty level of the policy network on the current observation, the high-level switch could monitor where the DRL agent may fail. Once the model uncertainty reaches a high level, conventional controllers, such as the APF \cite{warren1989global} controller, will take charge of the navigation task \cite{rana2020residual,Rana2020}. Although a high model uncertainty can well predict the poor performance of the corresponding DRL agent, it does not mean that a confident DRL agent can perform correctly. For example, a poorly trained DRL agent may crash into obstacles with high confidence in scenarios similar to its training scenario. 
\section{Preliminaries}\label{lim}
\subsection{Limitations of conventional LiDAR representations}\label{2issue}
\begin{figure}[t]
	\centering
	\includegraphics[width=0.50\linewidth]{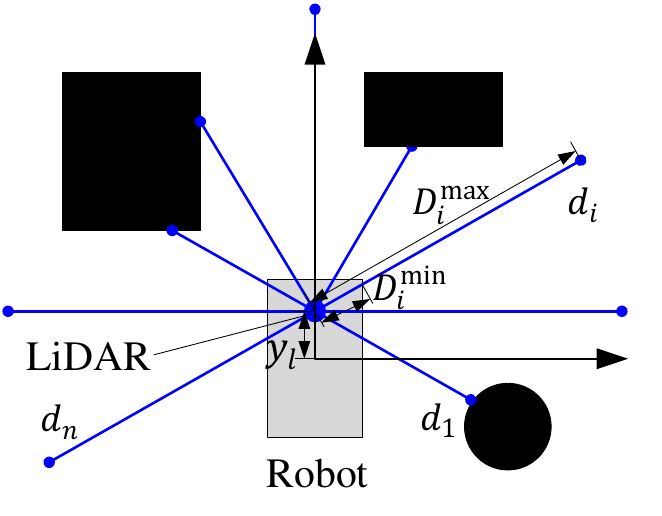}
	\caption{Illustration of LiDAR representation for DRL-based robot navigation.}
	\label{illu}
\end{figure}

The conventional representations of LiDAR readings are vectors containing raw laser scans or linearly normalized laser scans. Without loss of generality, as shown in Fig. \ref{illu}, we refer to raw LiDAR readings as $\mathbf{d}=[d_1,d_2,\ldots,d_n]$, where $d_i\in[D_i^{min},D_i^{max}]$ is the $i$-th measured distance value. $D_i^{max}$ denotes the maximum measuring distance of the LiDAR sensor, which is a property of the LiDAR sensor and is the same for all laser scans. $D_i^{min}$ denotes the minimum distance could be returned by the $i$-th laser beam, which is decided by the geometry shape of the robot and the installation position of the LiDAR. To reduce the input dimension, the raw laser scans are commonly compressed into \textit{m} values through 1D min-pooling \cite{Pfeiffer2018} as
\begin{equation}
\begin{aligned}
y_i=\min(d_{i\cdot k-k+1},d_{i\cdot k-k+2},\cdots,d_{i\cdot k}),
\end{aligned}
\end{equation}
where $k$ is the sampling window size ($k=1$ means min-pooling is not performed); $y_i\in[Y_i^{min},Y_i^{max}]$ is the \textit{i}-th min-pooled distance value ($Y_i^{min(max)}=\min(D_{i\cdot k-k+1}^{min(max)},D_{i\cdot k-k+2}^{min(max)},\cdots,D_{i\cdot k}^{min(max)})$). We refer to the min-pooled version as $\mathbf{y}=[y_1,y_2,\ldots,y_m]$ and the linearly normalized version of $\mathbf{y}$ as  $\mathcal{LN}(\mathbf{y})=[\mathcal{LN}(y_1),\mathcal{LN}(y_2),\ldots,\mathcal{LN}(y_m)]$, where $\mathcal{LN}(\cdot): \mathbb{R}\rightarrow\mathbb{R}$ denotes a linear mapping. When $y_i$ is close to $Y_i^{min}$, the robot is considered \hl{to be} too close to an obstacle, and the agent must focus on obstacle avoidance. We use a threshold $Y_i^T$ to determine whether the distance value $y_i$ is a short-distance value. If $y_i<Y_i^T$, the distance value $y_i$ is considered as a short-distance value, otherwise, it is considered as a long-distance value.

\begin{figure}[t]
    \centering
	  \subfloat[]{
       \includegraphics[width=0.4\linewidth]{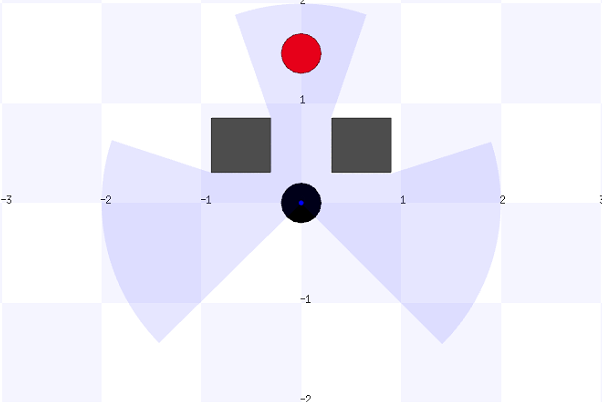}\label{limitation1}}
	  \subfloat[]{
        \includegraphics[width=0.4\linewidth]{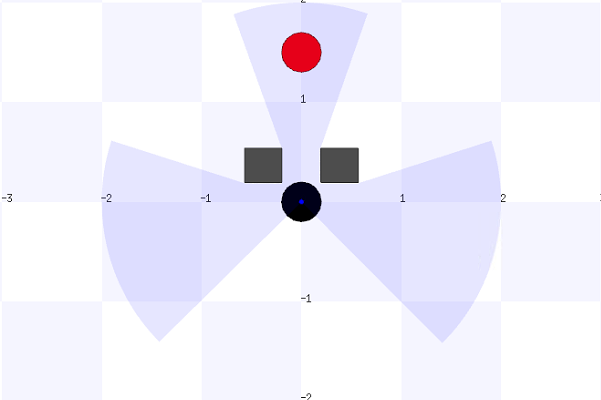}\label{limitation2}}
	\caption{Robot in two similar scenarios where the obstacle directions in both laser scans are the same. (a) The robot can move through the obstacle gate; (b) the robot cannot move through the obstacle gate.}
	\label{limitation}
\end{figure}

When learning navigation skills with $\mathbf{y}$ or $\mathcal{LN}(\mathbf{y})$ as the input, the DRL agent faces two challenges. First, the agent is required to generate completely different outputs on similar laser scans, which is challenging for DNNs. Since the short-distance values in LiDAR readings indicate that there are obstacles nearby, small changes in such values may lead to completely different navigation strategies. For example, as shown in Fig. \ref{limitation}, the laser scans in Fig. \ref{limitation1} are very similar to their counterparts in Fig. \ref{limitation2}. If one of the observations has been used to train the agent, the well-trained policy network is expected to output a similar policy on the other observation, because the two observations are similar. However, the navigation strategies for the two robots differ a lot\hl{. The} first robot should move through the obstacle gate, while the second one should bypass the gate. Similarly, when a robot navigates in a scenario more crowded than its training scenario, it may crash into obstacles based on the experience learned from a relatively open scenario. 

Second, the major proportion of input space of LiDAR readings is occupied by the long-distance values. Due to the high-dimensional observation space, it is nontrivial for the agent to collect training samples covering the entire observation space during training. Moreover, to learn effective obstacle-avoidance skills, crowded scenarios are usually preferred for training the agent \cite{Xie2018,Shi2020}. Since these scenarios are crowded, collected observations only contain a small fraction of the full observation space. When tested in relatively open scenarios, the agent may fail to complete navigation tasks because the received LiDAR observations are outside the training distribution. 
\subsection{Problem formulation}
\begin{figure}[t]
	\centering
	\includegraphics[width=0.4\linewidth]{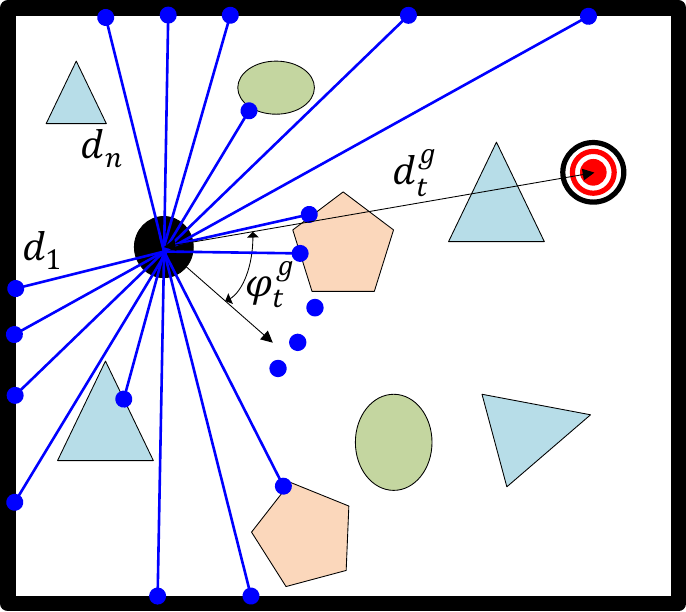}
	\caption{Illustration of the mapless robot navigation problem.}
	\label{problem}
\end{figure}

Due to the limitations of conventional input representations of laser scans, in this work, we aim to design a new input representation that can improve the generalization performance and speed up training of the DRL agent. Robot mapless navigation problem can be modelled as a sequential decision-making process. As shown in Fig. \ref{problem}, a robot, carrying a 2D LiDAR, is required to reach its goal location without colliding with obstacles. At step \textit{t}, the relative position of goal in robot coordinate frame $x_t^g=\{d_t^g,\varphi_t^g\}$ is given and assumed to be obtained by localization techniques, such as sound source localization \cite{Yook2016} or WIFI localization \cite{Chen2017}. We denote the input of the DRL agent as $x_t=\{\mathbf{p}_t,x_t^g,v_t,\omega_t\}$, where $\mathbf{p}_t$ is the new representation of LiDAR readings; $v_t$ and $\omega_t$ are current linear and angular velocities of the robot. The action $a_t$ of the robot is the velocity command. Given $x_t$, the agent takes $a_t$ under the current policy $\pi$. It then updates the next input $x_{t+1}$ based on new observations and receives a reward $r_t$. The reward function is as follows,
\begin{equation}
\begin{aligned}
r_t = 
\begin{cases}
r_{s}, & \text{if success,}\\
r_{c}, & \text{if crash,}\\
c_{1}\left(d_{t}^{g}-d_{t+1}^{g} \right), & \text{otherwise}.
\end{cases}
\end{aligned}
\end{equation}
where $c_1$ is a scaling constant. As shown, the reward function contains a positive part $r_s$ for encouraging success, a negative part $r_c$ for penalizing collision and a small dense part for encouraging the robot to move towards the target. The objective of this work is to find a representation $\mathbf{p}_t$ that can reduce the training steps and the $\mathbf{p}_t$-conditioned optimal policy $\pi^\ast$ that can maximize the discounted total rewards $G_t=\sum_{\tau=t}^{T}{\gamma^{\tau-t}r_\tau}$, where $\gamma\in\left[0,1\right)$ is a discount factor. \hl{In this paper, $c_1=2$, $r_s=10$, $r_c =-10$ and $\lambda=0.99$}.

\subsection{Soft actor critic}
In this paper, SAC is chosen as the DRL algorithm for learning navigation skills, because of its superior performance in DRL-based robot navigation  \cite{DeJesus2021}. It aims to maximize the entropy-regularized expected discounted cumulative reward. In SAC, The Q-function  \(Q^{ \pi } \left(x,a \right) \)  denotes the expected return from performing action  \(a\)  on input  \( x \) :
\begin{equation}
\begin{small}
\begin{aligned}
Q^{\pi}\left(x,a\right)=\mathbb{E}_{\pi}\left[G_{t=0}+\alpha\sum_{t=0}^{T}\gamma^{t}H\left(\pi \left(\cdot\vert x_{t}\right)\right)\vert x_{0}=x,a_{0}=a \right],
\end{aligned}
\end{small}
\end{equation}
where  \(H\left(\pi\left(\cdot\vert x_{t}\right)\right) =- \int _{\vert A \vert }^{} \pi  \left( a \vert x_{t} \right) \log  \pi  \left( a \vert x_{t} \right) da \)  denotes the entropy of the action;  the factor \(\alpha >0\) controls the contributions of the discounted cumulative reward and the entropy. SAC utilizes two critic (\textit{Q}) networks (parameterized by  \(  \phi _{1} \)  and  \(  \phi _{2} \)) to approximate the \textit{Q} value. As an off-policy DRL algorithm, it holds a replay buffer $\mathcal{B}$ to save the past state transitions $\left(x,a,r,x^\prime,d\right)$, where $d\in\{0,1\}$ is the termination signal. Sampling a mini-batch $\mathcal{M}$ from the replay buffer, the loss function $\mathcal{L}(\phi_{1,2})$ for the Q function is
\begin{equation}
\begin{small}
\begin{aligned}
&\mathcal{L}(\phi_{1,2})=\frac{1}{\left|\mathcal{M}\right|}\ \sum_{\left(x,a,r,x^\prime,d\right)\in\mathcal{M}}\left(Q\left(x,a|{\phi}_{1,2}\right)-\hat{Q}(x,a)\right)^2,\\
&\hat{Q}(x,a)=r+\gamma(1-d))(\min_{j=1,2}Q(x^\prime,\widetilde{a}^\prime|{\hat{\phi}}_{j})-\alpha\log\pi_{\theta}(\widetilde{a}^\prime\vert x^\prime)),
\end{aligned}
\end{small}
\end{equation}
where $\hat{Q}(x,a)$ denotes the target Q value, and $\widetilde{a}^\prime$ is sampled from $\pi\left(\cdot\vert x^\prime\right)$. The policy network is optimized to minimize the following loss function $\mathcal{L}(\theta)$ 
\begin{equation}
\begin{small}
\begin{aligned}
\mathcal{L}(\theta)=\frac{1}{\left|\mathcal{M}\right|}\ \sum_{x\in\mathcal{M}}\left(\alpha\log{\pi(\widetilde{a}\vert x,\theta)-\min_{j=1,2}Q(x,\widetilde{a}|{\phi}_{j})}\right).
\end{aligned}
\end{small}
\end{equation}During model training, the parameters of the critic and policy networks are updated to minimize $\mathcal{L}(\phi_{1,2})$ and $\mathcal{L}(\theta)$. \hl{In this paper, $\alpha=0.01$ and $\left|\mathcal{M}\right|=100$. }

\section{APPROACH}\label{approach}
\subsection{Input pre-processing of laser scans}\label{A1}
In this work, we focus on designing an efficient input representations of LiDAR readings that can alleviate the detrimental effects of the conventional representations. To achieve this, an element-wise IP function $\mathcal{P}_i\left(\cdot\right):\mathbb{R}\rightarrow\mathbb{R}$ is utilized to map each $y_i$ into $\mathcal{P}_i\left(y_i\right)$.  With threshold $Y_i^T$(see Section \ref{2issue}), we introduce a ratio $\rho^{PoS}\left(\mathcal{P}_i\left(y_i\right)\right)$ named \textit{PoS}  (proportion of 
short-distance range) to weigh the proportion of the short-distance range on the whole measuring range of $y_i$ in the space mapped by $\mathcal{P}_i\left(\cdot\right)$ as follows,
\begin{equation}
\begin{aligned}
\rho^{PoS}\left(\mathcal{P}_i\left(y_i\right)\right)=\left|\frac{\mathcal{P}_i\ (Y_i^T\ )-\mathcal{P}_i\ (Y_i^{min}\ )}{\mathcal{P}_i\ (Y_i^{max}\ )-\mathcal{P}_i\ (Y_i^{min}\ )}\right|.
\end{aligned}
\end{equation}
When no pre-processing or linear mapping $\mathcal{L}(\cdot)$ are performed, $\rho^{PoS}\left(y_i\right)=\rho^{PoS}\left(\mathcal{L}\left(y_i\right)\right)=\frac{Y_i^T-Y_i^{min}}{Y_i^{max}-Y_i^{min}}$. \textit{PoS} can well reflect two characteristics of the LiDAR input. \hl{Firstly}, a small \textit{PoS} value indicates the proportion of short-distance values in $\mathcal{P}_i\left(y_i\right)$ is small, which results in relatively little difference between two short-distance values and makes it difficult for DNNs to differentiate such a small difference. \hl{Secondly}, a small \textit{PoS} value indicates the range of long-distance values in $\mathcal{P}_i\left(y_i\right)$ is large, which suggests more training data containing various long-distance values are needed for maintaining the agent's performance in relatively open scenarios. Therefore, increasing the \textit{PoS} value can help address two issues of conventional input representations presented in Section \ref{2issue}. 

To increase the \textit{PoS} value of the distance representation, the following two conditions of $\mathcal{P}_i\left(\cdot\right)$ are proposed:
\begin{equation}\label{cond}
\begin{aligned}
    &\vert{\mathcal{P}_i\left(y_i\right)}\vert>0,\\
    &\mathcal{P}_i^\prime\left(y_i\right)\mathcal{P}_i^{\prime\prime}\left(y_i\right)<0.
\end{aligned}
\end{equation}
These two conditions ensure $\rho^{PoS}\left(\mathcal{P}_i\left(y_i\right)\right)>\rho^{PoS}\left(y_i\right)$, and the corresponding proof is as follows,
\begin{equation}
\begin{aligned}
\rho^{PoS}\left(\mathcal{P}_i\left(y_i\right)\right)&=\left|\frac{\mathcal{P}_i\left(Y_i^T\right)-\mathcal{P}_i\left(Y_i^{min}\right)}{\mathcal{P}_i\left(Y_i^{max}\right)-\mathcal{P}_i\left(Y_i^{min}\right)}\right|\\&=\frac{\left|\int_{Y_i^{min}}^{Y_i^T}{\mathcal{P}_i^\prime\left(\delta\right)}\,d\delta\right|}{\left|\int_{Y_i^{min}}^{Y_i^T}{\mathcal{P}_i^\prime\left(\delta\right)}\,d\delta\right|+\left|\int_{Y_i^T}^{Y_i^{max}}{\mathcal{P}_i^\prime\left(\delta\right)}\,d\delta\right|}\\&=\frac{1}{1+\frac{\left|\int_{Y_i^T}^{Y_i^{max}}{\mathcal{P}_i^\prime\left(\delta\right)}\,d\delta\right|}{\left|\int_{Y_i^{min}}^{Y_i^T}{\mathcal{P}_i^\prime\left(\delta\right)}\,d\delta\right|}}\\&>\frac{1}{1+\frac{\left|\mathcal{P}_i^\prime\left(Y_i^T\right)\right|\left(Y_i^{max}-Y_i^T\right)}{\left|\mathcal{P}_i^\prime\left(Y_i^T\right)\right|\left(Y_i^T-Y_i^{min}\right)}}\\&=\frac{\left|\mathcal{P}_i^\prime\left(Y_i^T\right)\right|\left(Y_i^T-Y_i^{min}\right)}{\left|\mathcal{P}_i^\prime\left(Y_i^T\right)\right|\left(Y_i^{max}-Y_i^{min}\right)}\\&=\rho^{PoS}\left(y_i\right).
\end{aligned}
\end{equation}
Since $\rho^{PoS}\left(\mathcal{P}_i\left(y_i\right)\right)>\rho^{PoS}\left(y_i\right)$, the proportion of long-distance values in the pre-processed representation $\mathcal{P}_i\left(y_i\right)$ is smaller than \hl{its} counterpart in the conventional representation of $y_i$. Accordingly, fewer training samples containing long-distance values are needed. Moreover, after mapped by $\mathcal{P}_i\left(\cdot\right)$, the differences between the pre-processed inputs have the following characteristics:
\begin{equation}
\begin{aligned}
\left|\mathcal{P}_i\left(y_{i,1}+\Delta y\right)-\mathcal{P}_i\left(y_{i,1}\right)\right|>\left|\mathcal{P}_i\left(y_{i,2}+\Delta y\right)-\mathcal{P}_i\left(y_{i,2}\right)\right|,
\end{aligned}
\end{equation}
where $y_{i,1}$ and $y_{i,2}$ ($y_{i,1}<y_{i,2}$) are two valid distance values in $y_i$, \hl{and} $\Delta y$ is the distance interval. As shown, after mapped by $\mathcal{P}_i\left(\cdot\right)$, the difference between two shorter distance values in $y_i$ is magnified relative to the difference between two longer distance values. Therefore, \hl{this characteristic} makes it easier for policy networks to learn different policies from the pre-processed representations of two LiDAR readings with similar small distance values.

Fortunately, many functions satisfy the two conditions presented in (\ref{cond}). In this work, three common functions, i.e., Exponential function $\mathcal{P}_i^E\left(\cdot\right)$, Logarithmic function $\mathcal{P}_i^L\left(\cdot\right)$ and Reciprocal function $\mathcal{P}_i^R\left(\cdot\right)$, are chosen as the IP functions and as follows,
\begin{equation}\label{IP_Exp}
\begin{aligned}
p_i^E=\mathcal{P}_i^E\ (y_i\ )=\lambda_i^{y_i},
\end{aligned}
\end{equation}\vspace{-2.5ex} 
\begin{equation}\label{IP_Log}
\begin{aligned}
&p_i^L=\mathcal{P}_i^L\ (y_i\ )=\ln\left(y_i-\eta_i\right),
\end{aligned}
\end{equation}\vspace{-2.5ex} 
\begin{equation}\label{IP_Rec}
\begin{aligned}
&p_i^R=\mathcal{P}_i^R\ (y_i\ )=\frac{1}{y_i-\beta_i},
\end{aligned}
\end{equation}
where $p_i^E$, $p_i^L$ and $p_i^R$ denote the pre-processed value of $y_i$; $\lambda_i\in\left(0,1\right)$ and $\eta_i,\beta_i\in\left(-\infty,Y_i^{min}\right)$ are parameters of these IP functions. We refer to $\mathbf{p}=\left\{p_1,p_2,\ldots,p_m\right\}$ as the pre-processed LiDAR input.

\subsection{Automating IP function adjustment}\label{AdaIP}
For $\mathcal{P}_i^E\left(\cdot\right)$, $\mathcal{P}_i^L\left(\cdot\right)$ and  $\mathcal{P}_i^R\left(\cdot\right)$, the suitable function parameters $\lambda_i$, $\eta_i$ and $\beta_i$ can be obtained by trial and error, but this is quite time-consuming. Instead of setting these parameters through trial and error, we can automate this procedure by treating those parameters as trainable variables and learning them through training. Without loss of generality, we refer to the IP function with trainable parameters as $\mathcal{P}_{\zeta_i}\left(\cdot\right)$, where $\zeta_i$ denotes trainable parameters. The parameters of the critic and policy networks are rewritten as $\bar{\theta}=\theta\cup\zeta_i$ and  ${\bar{\phi}}_{1,2}=\phi_{1,2}\cup\zeta_i$, and the gradient for updating the critic networks is
\begin{equation}\label{gcritic}
\begin{small}
\begin{aligned}
&\mathrm{\nabla}_{{\bar{\phi}}_{1,2}}\mathcal{L}(\bar{\phi})=\mathrm{\nabla}_{{\bar{\phi}}_{1,2}}\frac{1}{\left\vert \mathcal{M}\right|}\ \sum_{\left(x,a,r,x^\prime,d\right)\in\mathcal{M}}\left(Q\left(x,a\vert {\bar{\phi}}_{1,2}\right)-\hat{Q}_(x,a)\right)^2.
\end{aligned}
\end{small}
\end{equation}
Besides, the gradient for updating the policy network is

\begin{equation}
\label{gpolicy}
\begin{small}
\begin{aligned}
\mathrm{\nabla}_{\bar{\theta}}\mathcal{L}(\bar{\theta})=\mathrm{\nabla}_{\bar{\theta}}\frac{1}{\left|\mathcal{M}\right|}\ \sum_{x\in\mathcal{M}}\left(\alpha\log{\pi(\widetilde{a}\vert x,\bar{\theta})-\min_{j=1,2}Q(x,\widetilde{a}|{\bar{\phi}_{j})}}\right).
\end{aligned}
\end{small}
\end{equation}
Using (\ref{gcritic}) and (\ref{gpolicy}), the optimal parameters of these IP functions can be learned simultaneously with the optimal policy. 

\subsection{Neural network models}
\begin{figure*}[t]
    \centering
	  \subfloat[Model\_0 \cite{Pfeiffer2018}]{
       \centering\includegraphics[width=0.22\linewidth]{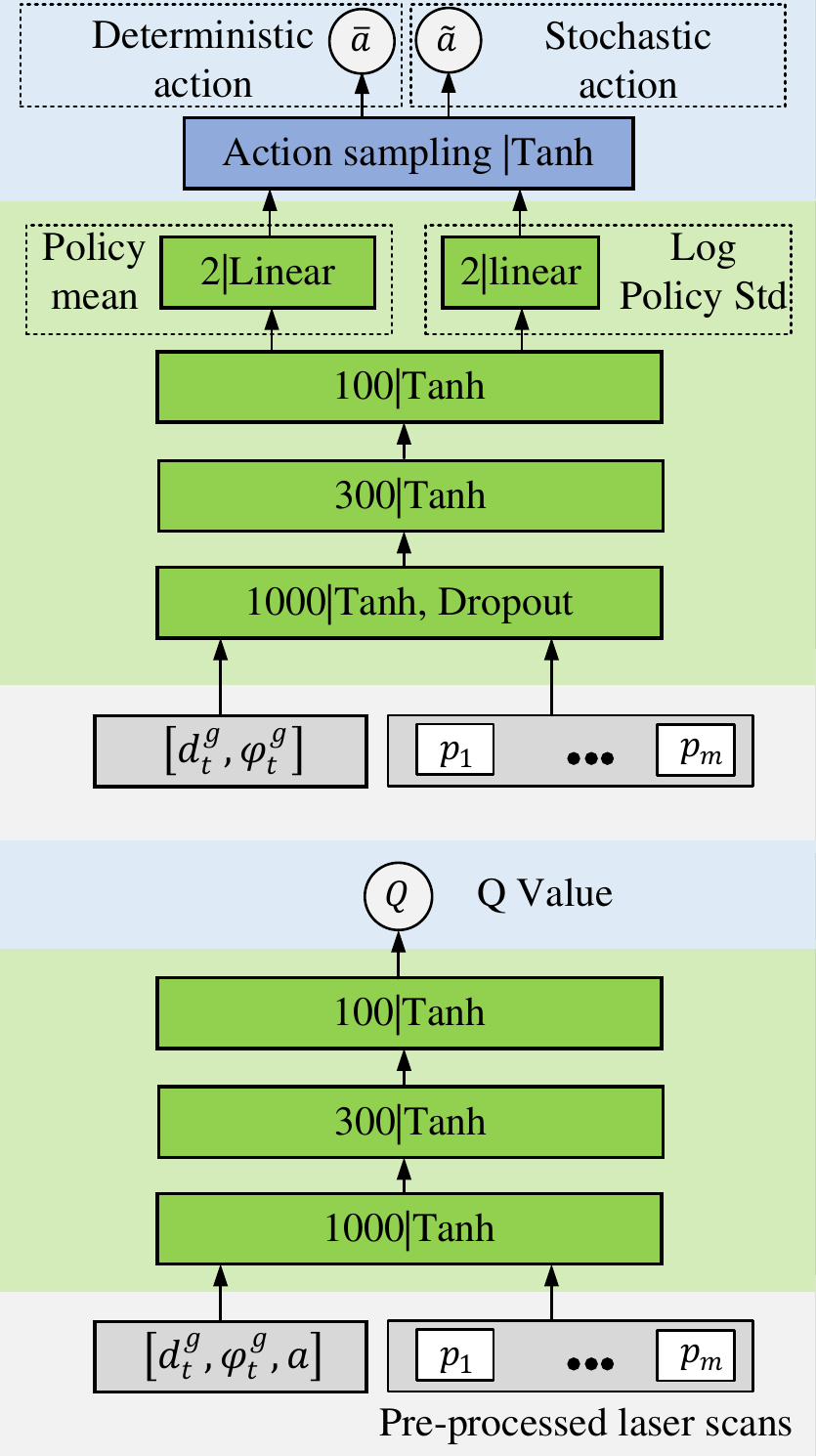}\label{net1}}
       \hfill
	  \subfloat[Model\_1 \cite{Tai2017}]{
        \centering\includegraphics[width=0.22\linewidth]{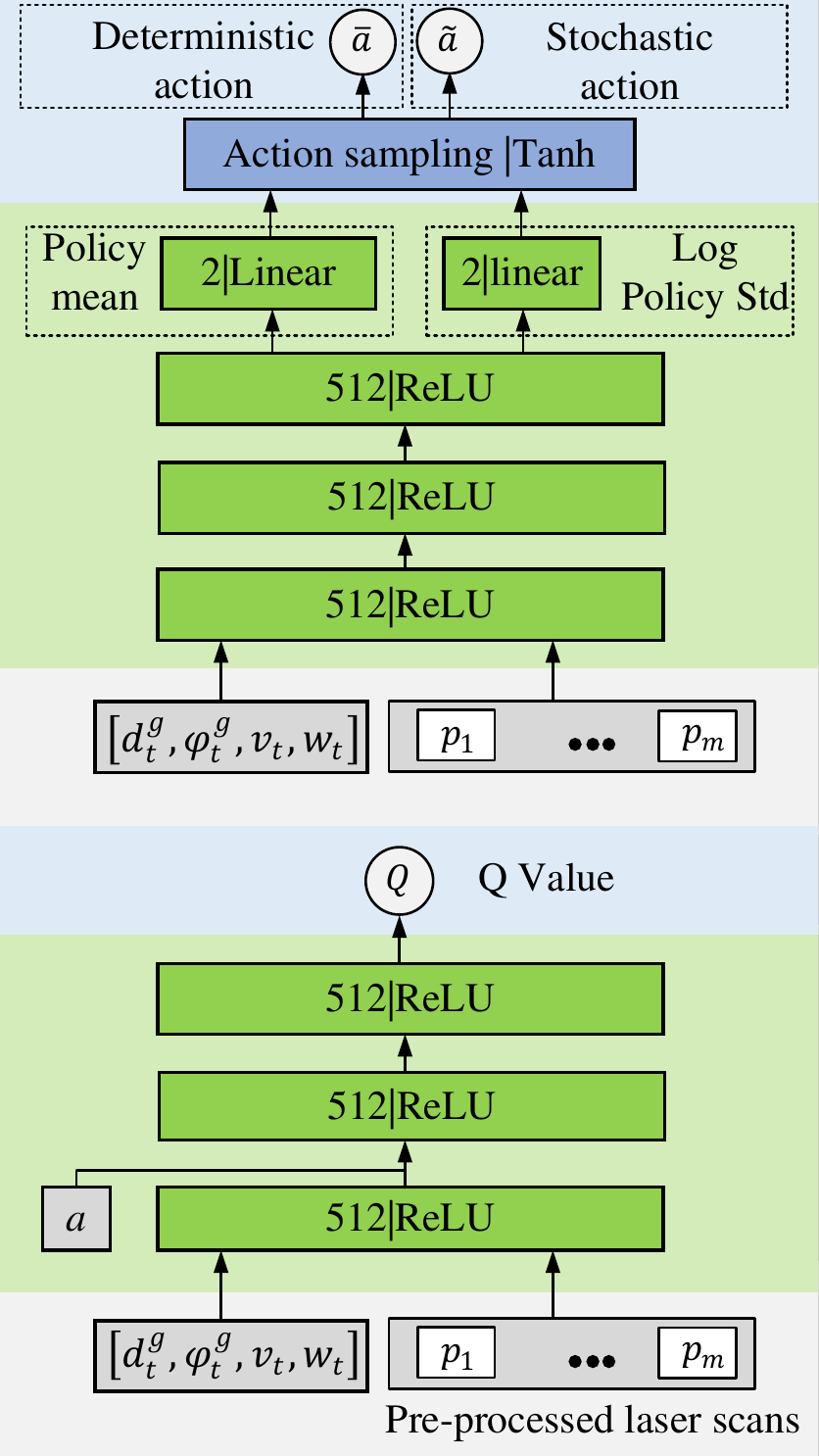}\label{net2}}
        \hfill
	  \subfloat[Model\_2 \cite{Fan2020DistributedScenarios}]{
        \centering\includegraphics[width=0.22\linewidth]{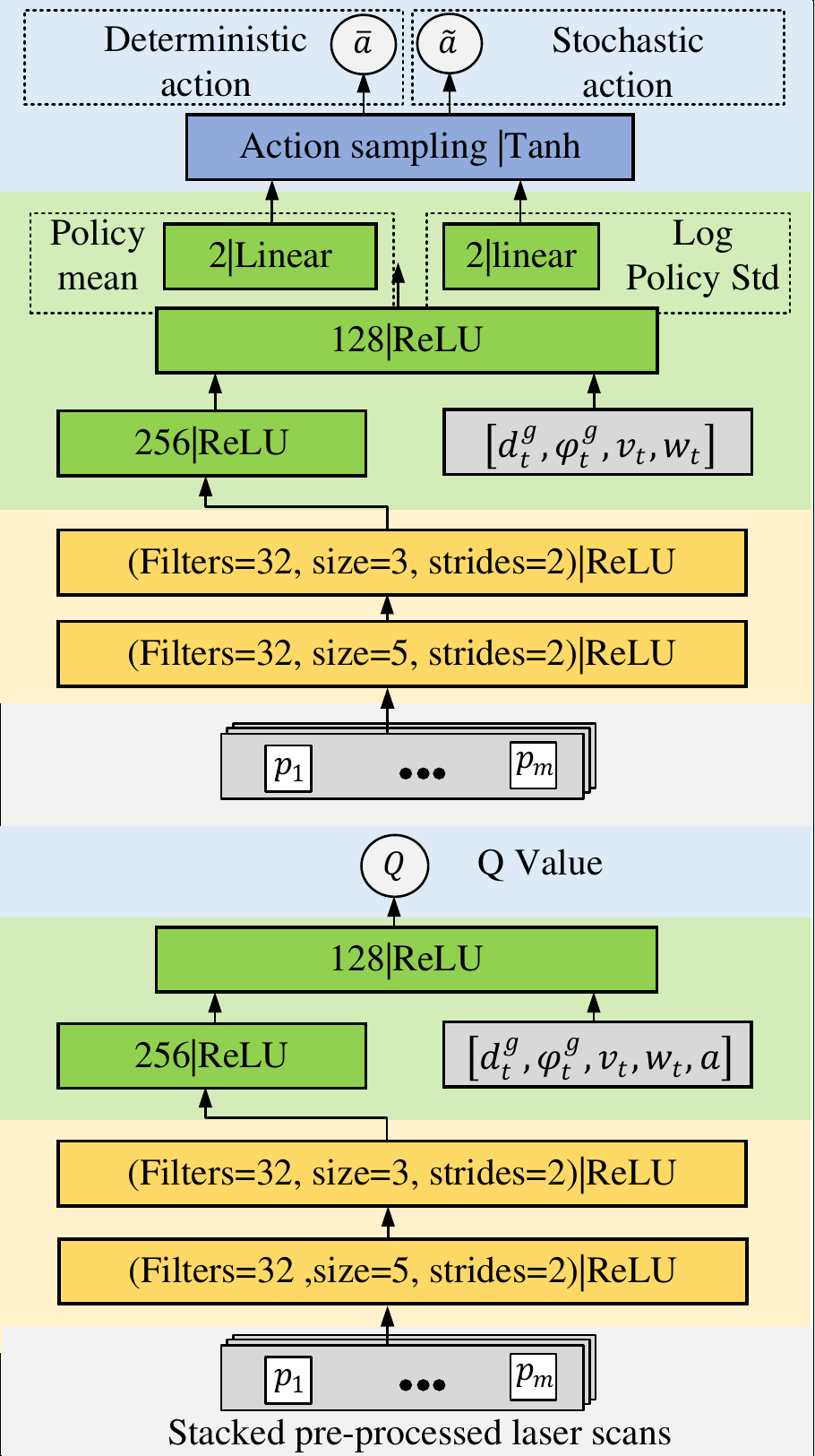}\label{net3}}
        \hfill
	  \subfloat[Model\_3]{
        \centering\includegraphics[width=0.22\linewidth]{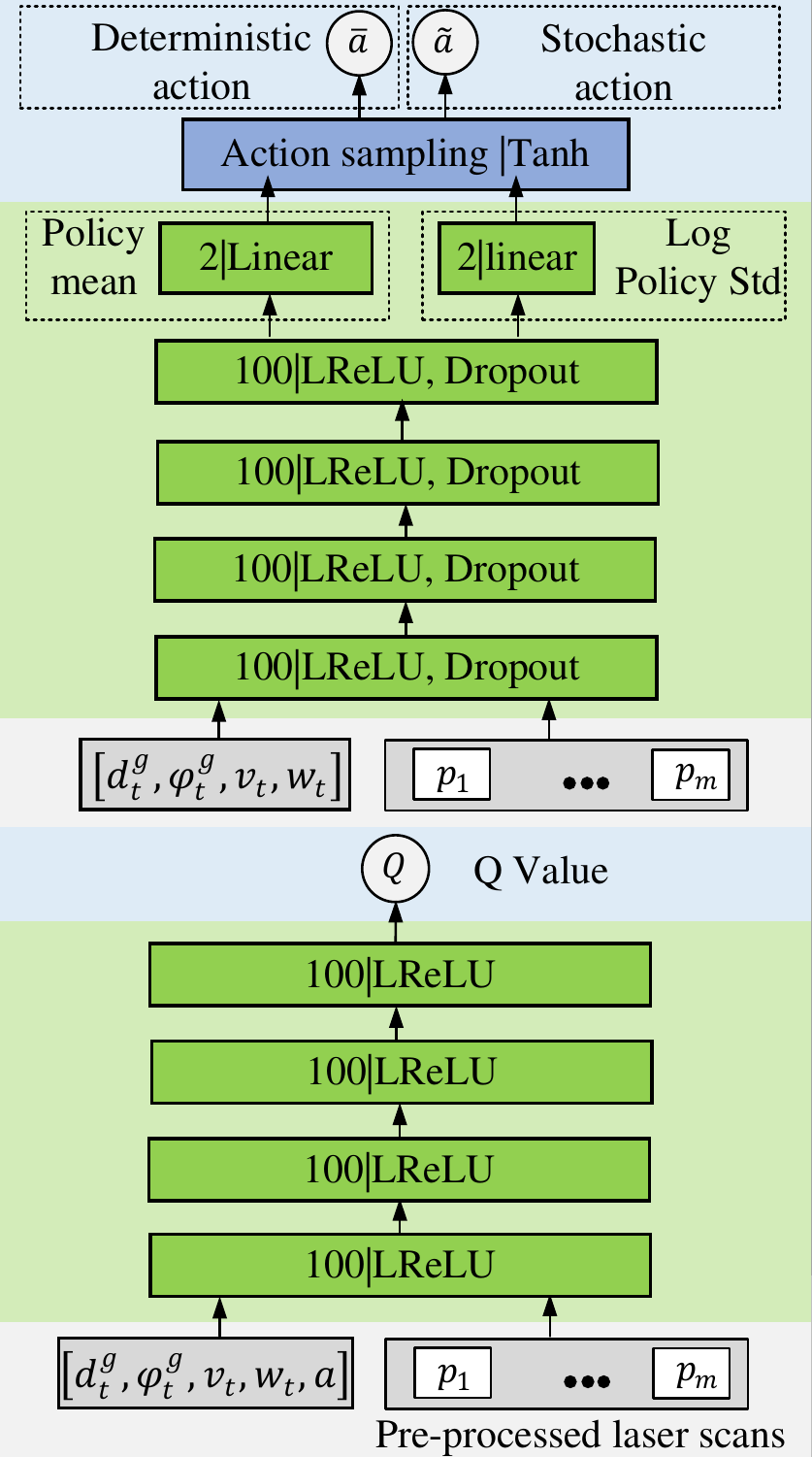}\label{net4}}
	\caption{Neural network architectures used for evaluating the IP approaches.}
	\label{nets}
\end{figure*}

To assess the effect of the proposed IP approach on different neural network structures, three public neural network models shown in Fig. \ref{nets}, namely Model\_0 \cite{Pfeiffer2018}, Model\_1 \cite{Tai2017}, Model\_2 \cite{Fan2020DistributedScenarios}, \hl{were} selected for training. As shown, these three models contain two fully-connected (FC) Model and one convolutional-neural-network (CNN) model. Besides, a four-layer FC model (Model\_3, see Fig. \ref{net4}) with leaky rectified linear unit (LReLU) as activation function is presented. Its policy is regularized by Dropout, which helps to enhance the generalization performance of the DRL agent \cite{Liu2019}. With the specific network structure and SAC algorithm, the DRL agent can learn to navigate in simulation, and its training procedure is given in Algorithm \ref{algorithm}.
\begin{algorithm}[t]
\label{algorithm}
\SetAlgoLined
 Initialize parameters of policy network (with trainable IP parameters) $\bar{\theta}$,  parameters Q-value function (with trainable IP parameters) $\bar{\phi}_{1}$, $\bar{\phi}_{2}$, empty replay buffer $\mathcal{B}$ \;
 \For{\upshape{episode}$=1$, 2, \ldots, }
 {  Reset the training environment and initialize $t=0$\;
      Obtain initial observation $x_0$\;   
 \While{$t<T_{max}$ \upshape{\textbf{and} not terminate}}
   {
      \eIf{\upshape{training}}{Sample action $a_t\sim\pi_{\theta}(x_t)$\;}
      {Sample a random action $a_t$\;}
      Execute $a_t$ in simulation\;
      Obtain next observation $x_{t+1}$, reward $r_t$ and the termination signal $d_t$\;
      Store $\{x_t,a_t,r_t,x_{t+1},d_t\}$ in $\mathcal{B}$\;
      $t\leftarrow t+1$, $x_t\leftarrow x_{t+1}$\;
      \If{\upshape{training}}{   
      Sample a mini-batch $\mathcal{M}$ from $\mathcal{B}$\;
      	Update $\bar{\phi}_{1}$, $\bar{\phi}_{2}$ using (\ref{gcritic})\;
        Update $\bar{\theta}$ using (\ref{gpolicy})\;
      }
      }}
\caption{learn to navigate with adaptively parametric IP}
\end{algorithm}

\section{Implementation and Tests in Simulation}\label{sim}
In this section, we \hl{evaluated} the proposed IP approaches using three case studies. \hl{Firstly}, we \hl{trained} a circular robot to navigate in a relatively open scenario. In this case, the LiDAR \hl{was} mounted on the top center of the robot, and hence all distance values \hl{shared} the same IP function. \hl{Secondly}, we \hl{trained} a rectangular robot to navigate in a complex scenario. In this case, the IP function is unique for the distance value measured from each direction, which is more common in real-world applications. Finally, we \hl{trained} a circular robot to navigate in a challenging scenario, and the learned navigation controller \hl{would} be directly deployed to a real robot for real-world testing. The descriptions of all the IP functions used in the following case studies are summarized in Table \ref{tab:table-0}. \hl{The learning rates for actor and critic networks were both $10^{-4}$; the initial value of IPAPRec trainable parameter $\beta_i=0$; the number of episodes for choosing random action was 100.} 

\hl{The DRL agent for the mobile robot was trained in simulation environments built on ROS Stage {\cite{Stage_rosWiki}}, a commonly-used simulator for mobile robots. Each training scenario was generated using a 2D map, with black areas representing areas occupied by obstacles. In the simulated scenario, a robot model was built, and the collision checker constantly monitored the robot for collisions with obstacles. At each time step, the DRL agent received the laser scans from the 2D LiDAR, and the DNN controller output the control command, i.e., the linear and angular velocities, to the robot. Afterwards, the robot performed the command and received a reward based on the reward function. The control frequency of the robot was 5Hz, and the maximum time steps per episode $T_{max}$ was 200.}

\begin{table}
\caption{\label{tab:table-0} The list of IP functions.} 
\renewcommand\arraystretch{1.2}
\centering
% \usepackage{multirow}
\setlength{\tabcolsep}{1.4mm}{\begin{tabular}{ll}
\hline
\hline
IP Methods   & Descriptions                                                             \\ \hline
Raw              & Raw laser scans                                                         \\
LNorm            & Linearly normlized into [0, 1], see (\ref{LNorm})                       \\
IPAPExp            & IP with adaptively parametric exponential function, see (\ref{IP_Exp})                   \\
IPAPLog            & IP with adaptively parametric logarithmic function, see (\ref{IP_Log})                   \\
IPAPRec          & IP with adaptively parametric reciprocal function, see   (\ref{IP_Rec}) \\
IPAPRecN         & Normalized version of IPAPRec, see (\ref{IPAPRecN})                  \\ \hline
\hline
\end{tabular}}
\end{table}

\subsection{Case Study 1: training a circular robot to navigate}
\begin{figure}[t]
	\centering
	\includegraphics[width=0.3\linewidth]{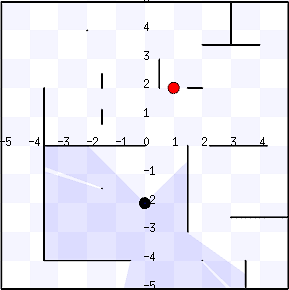}
	\caption{Env\_0 \cite{Pfeiffer2018}: the simulated scenario used in Case Study 1.}
	\label{Env0}
\end{figure}

In the first case study, we \hl{evaluated} our IP approaches by training a circular robot to navigate. As shown in Fig. \ref{Env0}, the training scenario Env\_0 \cite{Pfeiffer2018} \hl{was} a $10\times10\text{m}^2$ room, which \hl{contained} multiple thin obstacles. In this room, a simulated differential robot (the black circle), with a radius of $0.2 \text{m}$, \hl{learned} how to reach its target (the red circle). Its maximum linear and angular velocities \hl{were} $0.5\text{m/s}$ and $\frac{\pi}{2}\text{rad/s}$, respectively. It \hl{carried} a 2D LiDAR on the top center for obstacle detection. The LiDAR  \hl{had} a FOV, angular resolution and scanning range of $270^\circ$, $0.25^\circ$ (1080 laser beams) and $30\text{m}$, respectively. For the DRL agent, its DNN model \hl{was} Model\_0 given in Fig. \ref{net1}, and the input size of laser scans \hl{was} reduced to 36 through min-pooling. 

\begin{figure}[t]
	\centering
	\includegraphics[width=0.8\linewidth]{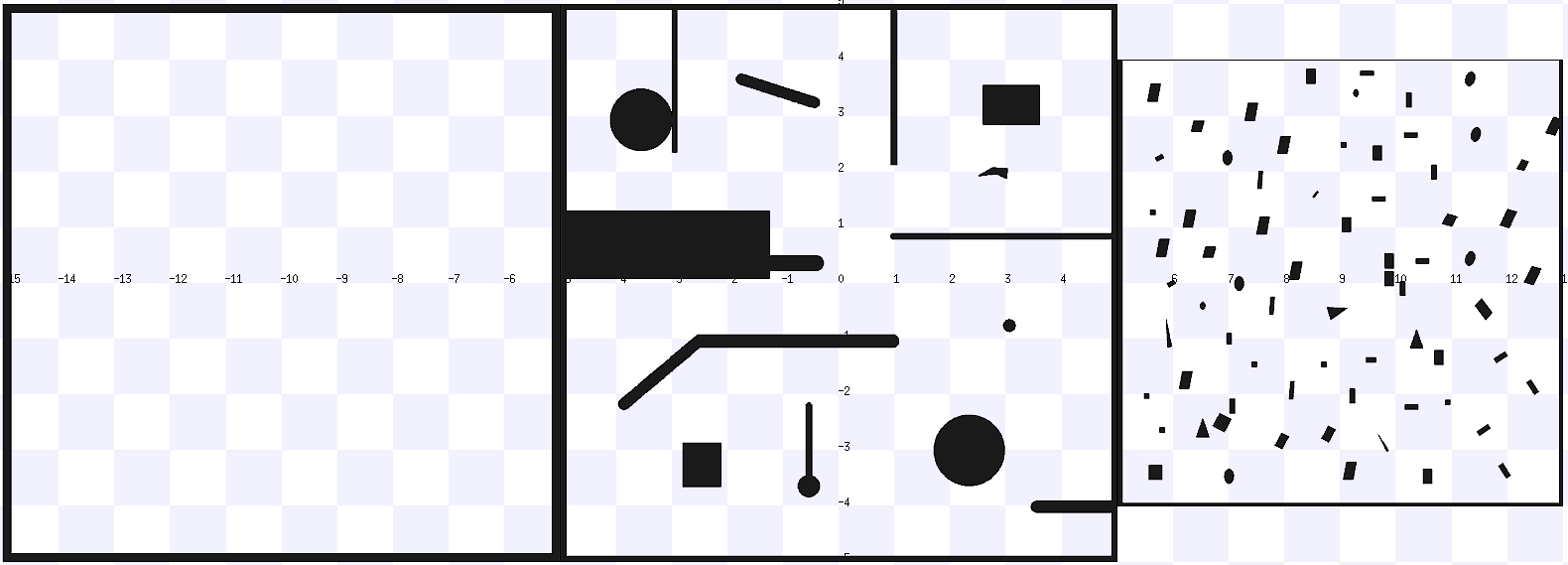}
	\caption{The simulated scenarios used for testing, namely Env\_1, Env\_2 \cite{Pfeiffer2018}, and Env\_3 \cite{zhang2021learn} (left to right).}
	\label{Env1}
\end{figure}

At the beginning of each episode, the agent \hl{started} from a randomly chosen obstacle-free location. Each episode \hl{terminated} once the robot \hl{reached} its goal, \hl{hit} obstacles, or \hl{timed} out. For evaluation, every 5000 training steps, the agent \hl{was} required to perform 50 different tasks in Env\_0 and in each unseen testing scenario, namely Env\_1, Env\_2, and Env\_3 (see Fig. \ref{Env1}), respectively. Env\_1 \hl{was} an empty room, Env\_2 \hl{was} an office-like room and Env\_3 \hl{was} a room with high obstacle density. These three scenarios \hl{differed} widely in the degree of \hl{crowdedness}. Env\_1 \hl{was} chosen to investigate whether the agent can choose near-optimal paths during the goal-reaching process, while Env\_2 and Env\_3 \hl{were} used to test the obstacle-avoidance performance of the DRL agent. The number of total training steps and replay buffer size \hl{were} both $10^5$.

Three adaptively parametric IP functions, i.e., IPAPExp, IPAPLog and IPAPRec (see Table \ref{tab:table-0}), \hl{were} implemented to pre-process the LIDAR input. For comparison, two conventional IP methods named Raw and LNorm \hl{were} also implemented. LNorm \cite{Tai2017} linearly \hl{compressed} the distance range into $[0,1]$ as follows,
\begin{equation}\label{LNorm}
{\mathcal{LN}}\left(y_i\right)=\frac{y_i}{Y_i^{max}}.
\end{equation}
In addition, a normalized version of IPAPRec, namely IPAPRecN, \hl{was} implemented as follows,
\begin{equation}\label{IPAPRecN}
\mathcal{P}^R_N\left(y_i\right)=\frac{Y_{i}^{min}-\beta_i}{y_{i}-\beta_i}.
\end{equation}

\begin{figure}[t!]
	\centering
	\includegraphics[width=0.98\linewidth]{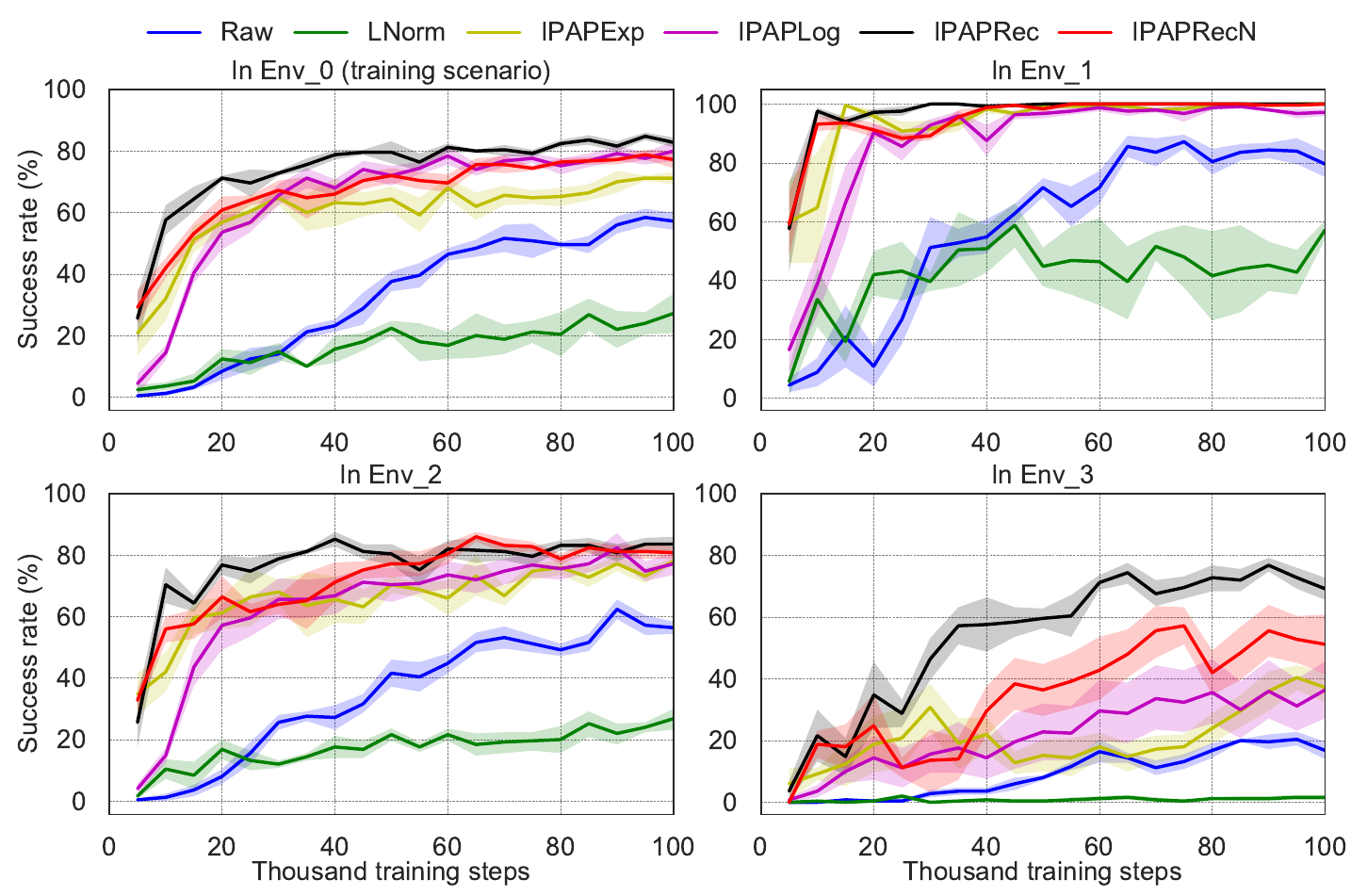}
	\caption{Learning curves of DRL agents trained in Env\_0 with Model\_0 using different IP functions.}
	\label{LC_1}
\end{figure}

Each model \hl{was} trained five times using different random seeds to assess the stability of the learning algorithm. The metric used for evaluation \hl{was} the success rate of the agent in 50 runs in each scenario, and the learning curves \hl{were} plotted in Fig. \ref{LC_1}. As shown, all the proposed IP approaches \hl{outperformed} the two conventional approaches in terms of success rate. Among the three IP approaches, IPAPRec \hl{performed} the best in all scenarios: it \hl{achieved} the highest success rate and \hl{learned} the fastest. Hence, we \hl{chose} IPAPRec as our default IP approach in the following two cases studies.

\subsection{Case Study 2: training a rectangular robot to navigate}

\begin{figure}[t!]
	\centering
	\includegraphics[width=0.3\linewidth]{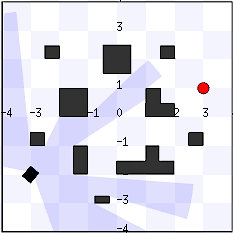}
	\caption{Env\_4 \cite{xie2021}: the simulated scenario used in Case Study 2.}
	\label{Env2}
\end{figure}

In this case study, we \hl{evaluated} the performance of our IPAPRec/IPAPRecN approaches in a more general case, i.e., training a rectangular robot to navigate. As the robot \hl{was} rectangular and the LiDAR \hl{was} mounted on the top front of the robot ($y_l=0.1$m, see Fig. \ref{illu}), the minimum measuring range $D_{i}^{min}$ and its min-pooled value $Y^{min}_i$ \hl{varied} with the direction index $i$ . Hence, the parameter $\zeta_i$ of each IP function $\mathcal{P}_{\zeta_i}\left(\cdot\right)$ \hl{was} unique for each $y_i$ in this case.  As shown in Fig. \ref{Env2}, the training scenario \hl{was} Env\_4 \cite{xie2021}, and the robot \hl{had} the same size as the Pioneer3-DX robot, with a size of $0.455\times0.381\text{m}^2$. The equipped LiDAR \hl{had} a FOV, angular resolution and scanning range of $240^\circ$, $0.47^\circ$ (512 laser beams) and $5.6\text{m}$, respectively. The speed limits of the robot \hl{remained} the same as in the first case. The adopted DNN model \hl{was} Model\_1 (see Fig. \ref{net2}), where the input size of laser scans \hl{was} reduced to 32 by min-pooling. With this model, four IP approaches, i.e., Raw, LNorm, IPAPRec and IPAPRecN, \hl{were} implemented and evaluated. 

\begin{figure}[t!]
	\centering
	\includegraphics[width=0.98\linewidth]{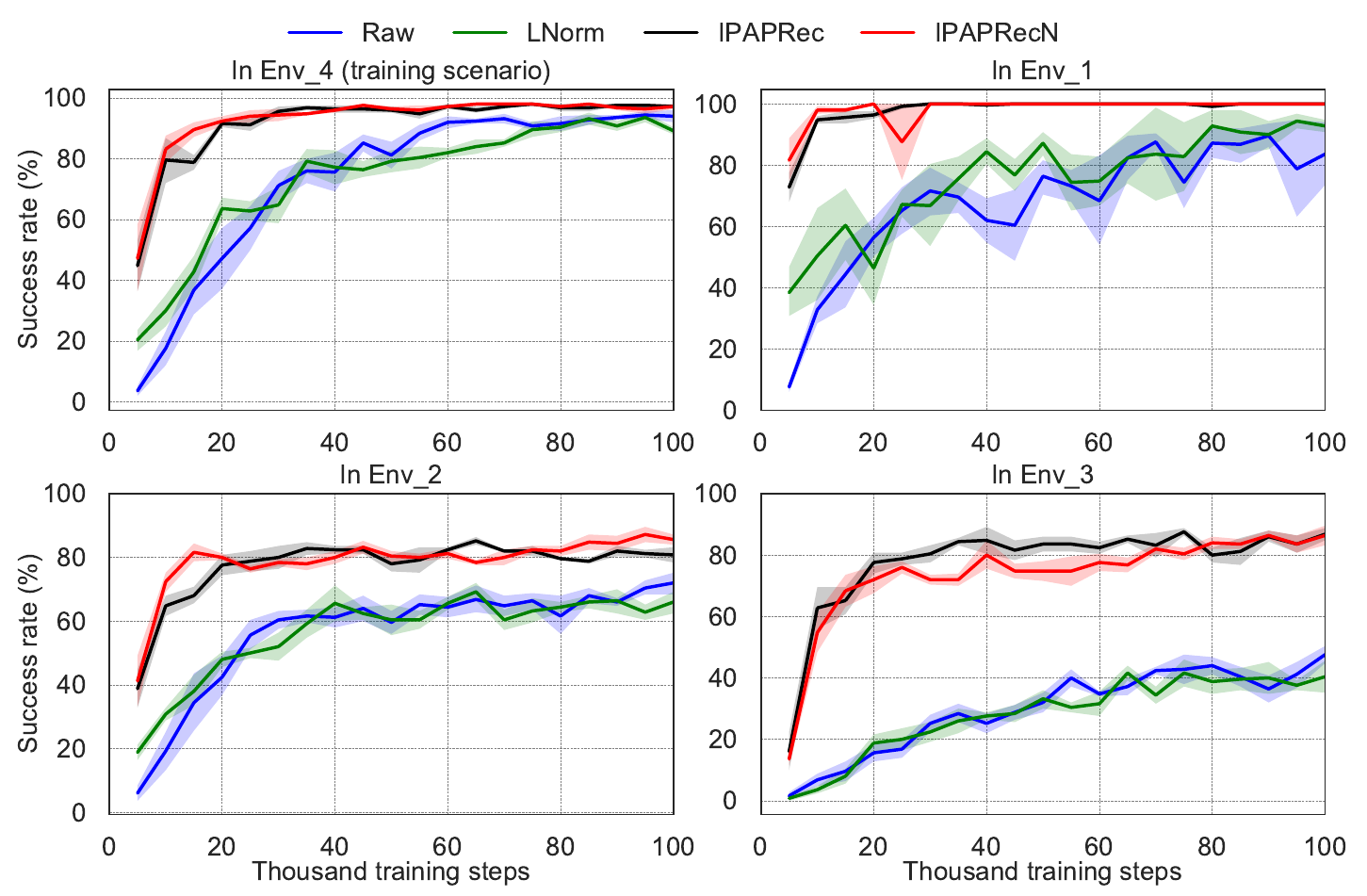}
	\caption{Learning curves of DRL agents trained in Env\_4 with Model\_1 using different IP functions.}
	\label{LC_2}
\end{figure}

The training and testing details \hl{were} the same as in the first case, except the room size of Env\_3 \hl{was} increased into $10\times10\text{m}^2$ due to the increased size of the new robot. As shown in Fig. \ref{LC_2}, at the end of training, all approaches \hl{achieved} a success rate close to 100\% in the training scenario. However, in the testing scenarios, only our IPAPRec and IPAPRecN approaches \hl{maintained} the high performance.

\begin{figure}[t!]
	\centering
	\includegraphics[width=0.98\linewidth]{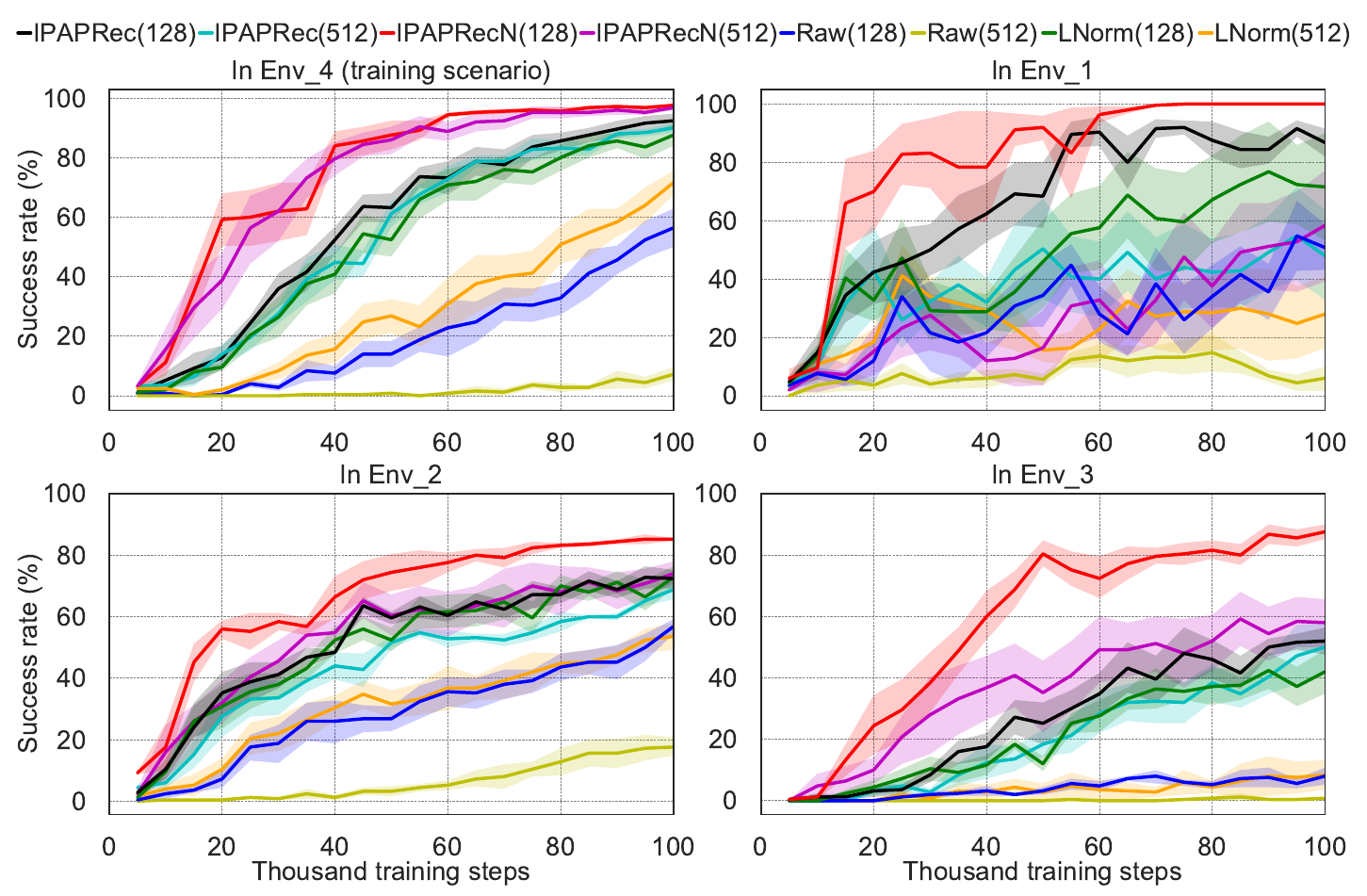}
	\caption{Learning curves of DRL agents trained in Env\_4 with Model\_2 using different IP functions.}
	\label{LC_3}
\end{figure}

As 1D CNN structures \hl{were} also commonly used in DRL-based navigation \cite{Xie2018,Fan2020DistributedScenarios}, we also \hl{investigated} the effect of our IPAPRec/IPAPRecN approaches on 1D CNN models. Following \cite{Fan2020DistributedScenarios}, the adopted 1D CNN model \hl{was} Model\_2 (see Fig. \ref{net3}). Its LiDAR input \hl{was} stacked laser scans received from \hl{the} last three measures with a format of $[3,m]$, where $m$ is the number of laser beams. For CNN models, the value of $m$ is usually greater than their counterparts in FC models. Hence, in this case, $m$ \hl{was} set as 512 (without min-pooling, the same as \cite{Fan2020DistributedScenarios}) and 128 (a min-pooled version), respectively. The training and testing details \hl{were} the same as for Model\_1. As shown in Fig. \ref{LC_3}, for CNN models, our IPAPRecN (128) approach \hl{performed} the best, which \hl{was} the only one that \hl{achieved} 100\% success rate in Env\_1 and more than 80\% success rate in Env\_2 and Env\_3. 
\subsection{Case Study 3: training a robot to navigate in a complex scenario}\label{case3}
\begin{figure}[t!]
	\centering
	\includegraphics[width=0.3\linewidth]{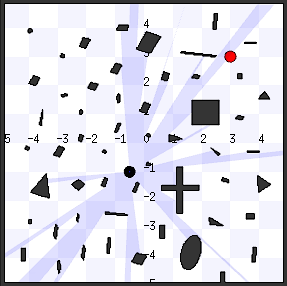}
	\caption{Env\_5: the simulated scenario used in Case Study 3.}
	\label{Env3}
\end{figure}

In this case study, we \hl{trained} a DRL agent with the capability to navigate in real-world crowded scenarios. The robot and LiDAR used \hl{were} the same as in Case Study 1. Due to the relatively simple training scenario in Case Study 1, the DRL agent \hl{failed} to collect enough samples covering the crowded situation and \hl{performed} relatively poorly in unseen crowded scenarios such as Env\_3. To improve the agent's performance in crowded scenarios, we \hl{designed} a challenging training scenario, i.e., Env\_5 (see Fig. \ref{Env3}), which \hl{was} much more crowded than Env\_0. Besides, to help enhance the agent's generalization capability, we \hl{used} the four-layer FC model, i.e., Model\_3, with dropout-based policy regularization \cite{Liu2019}. For this model, the input size of laser scans \hl{was} reduced to 30 through min-pooling. With this model, four IP approaches, i.e., Raw, LNorm, IPAPRec and IPAPRecN, \hl{were} implemented and evaluated. \hl{In addition, we compared them with two traditional local-planning methods, namely DWA and APF.}
\begin{figure}[t!]
	\centering
	\includegraphics[width=0.98\linewidth]{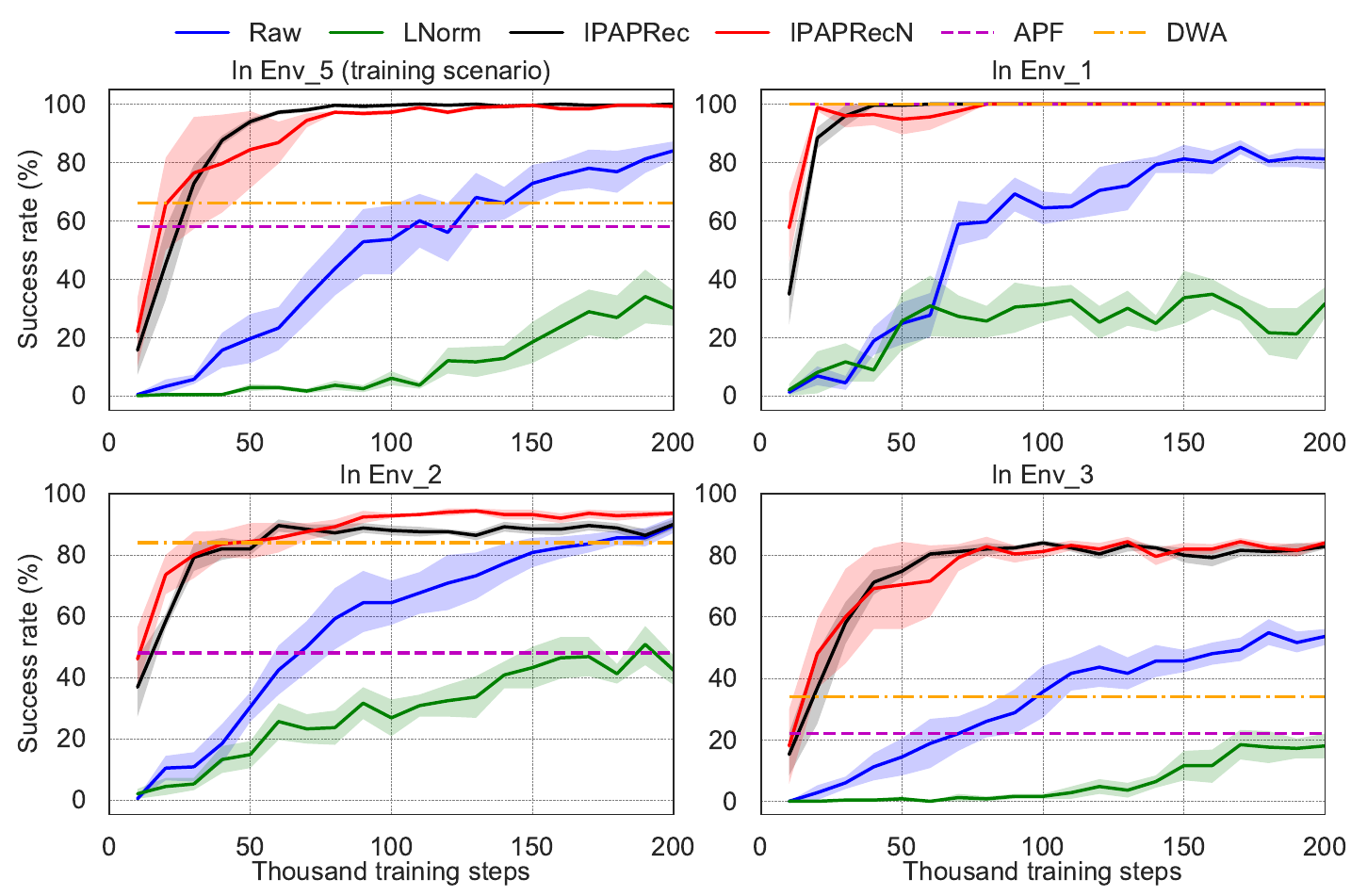}
	\caption{\hl{Learning curves of DRL agents trained in Env\_5 with Model\_3  using different IP functions, DWA and APF.}}
	\label{LC_4}
\end{figure}

\begin{figure}[t!]
    \centering
	  \subfloat[Raw in Env\_1]{
       \centering\includegraphics[width=0.36\linewidth]{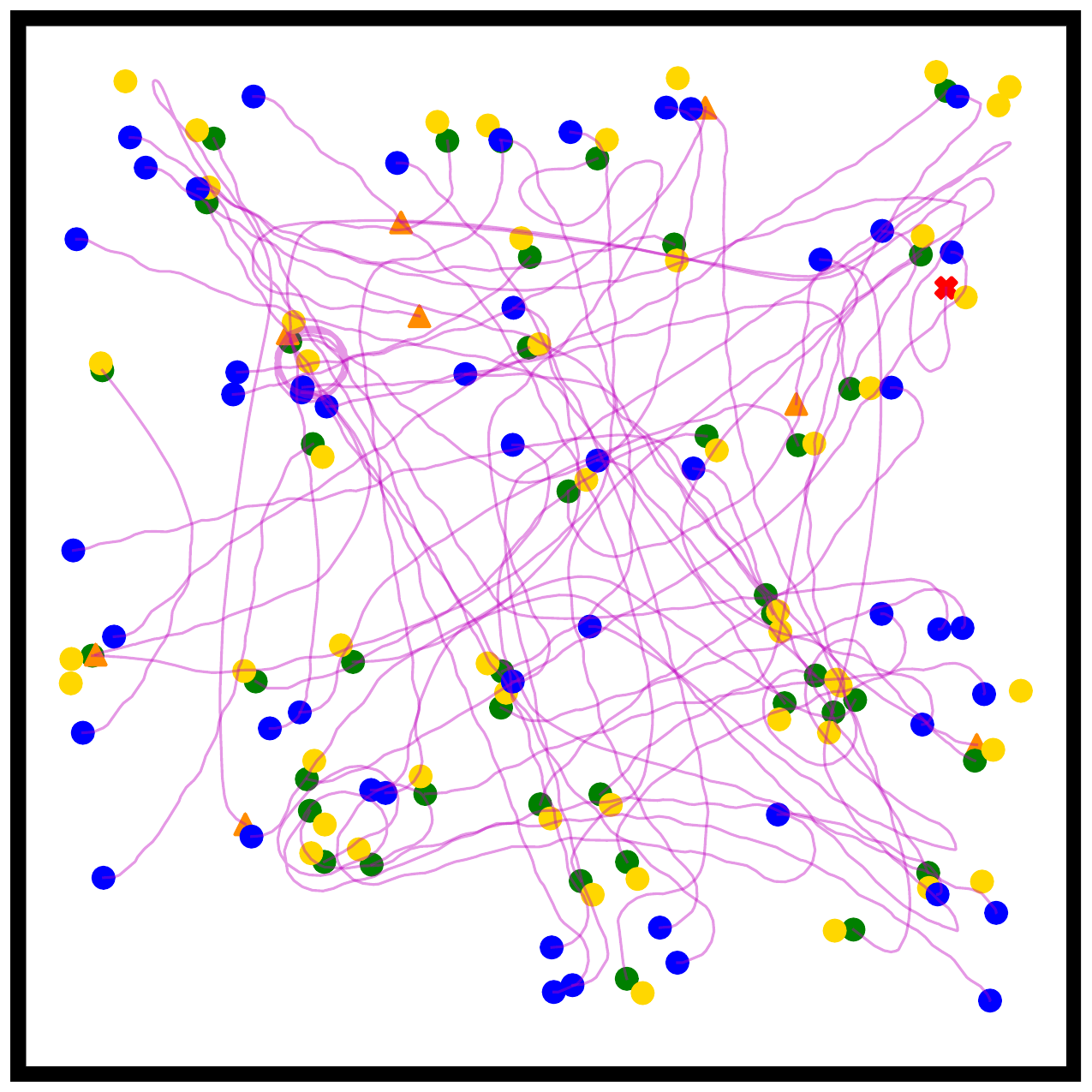}\label{traj1}}
	  \subfloat[IPAPRec in Env\_1]{
        \centering\includegraphics[width=0.36\linewidth]{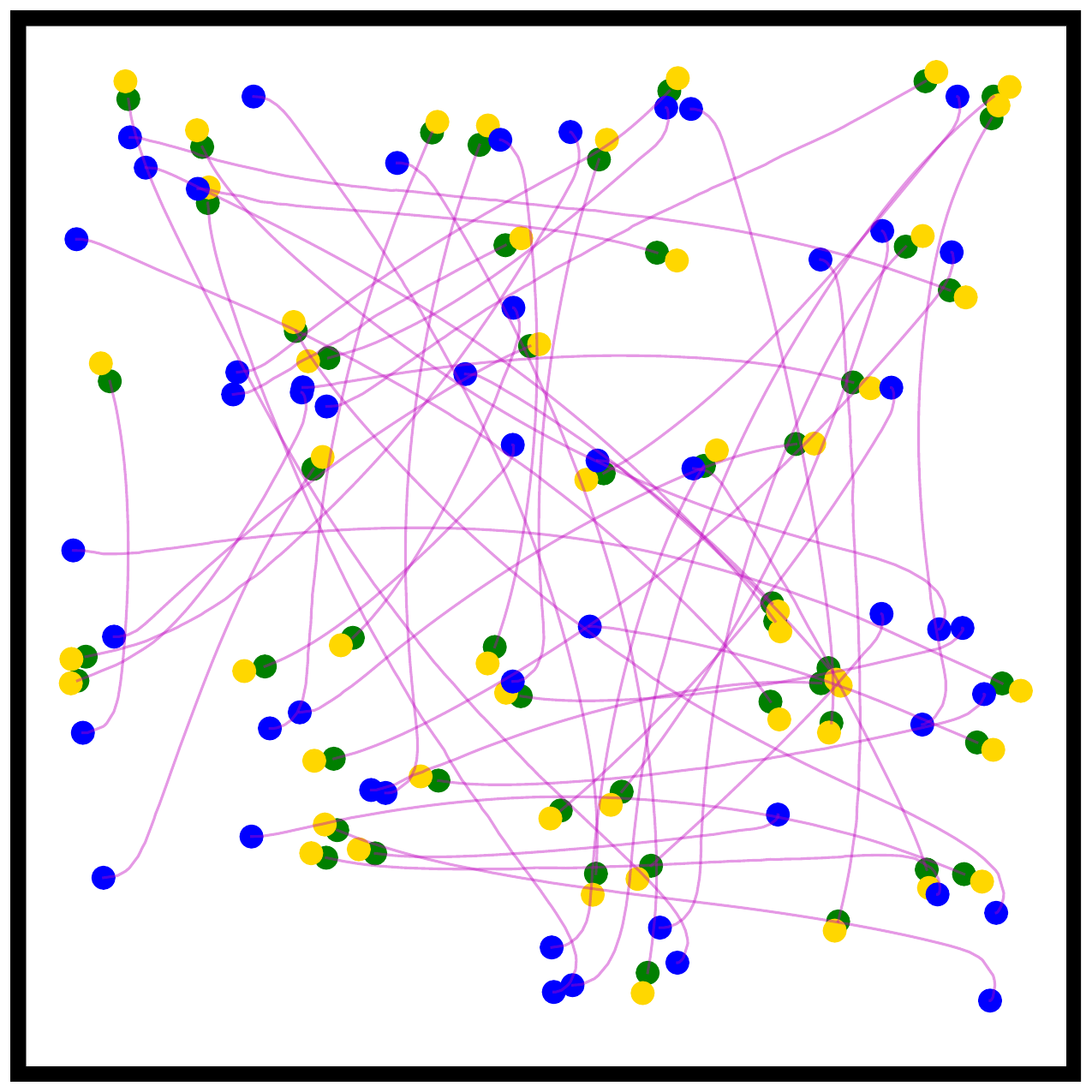}\label{traj2}}
	\caption{Trajectories of agents trained with Raw and IPAPRec tested in scenario Env\_1. The blue and yellow circles represent the starting point and goal point, respectively. The success and timeout termination states are marked with a green circle and an orange triangle, respectively. }
	\label{traj}
\end{figure}

\begin{figure}[t]
    \centering
	  \subfloat[]{
       \includegraphics[width=0.40\linewidth]{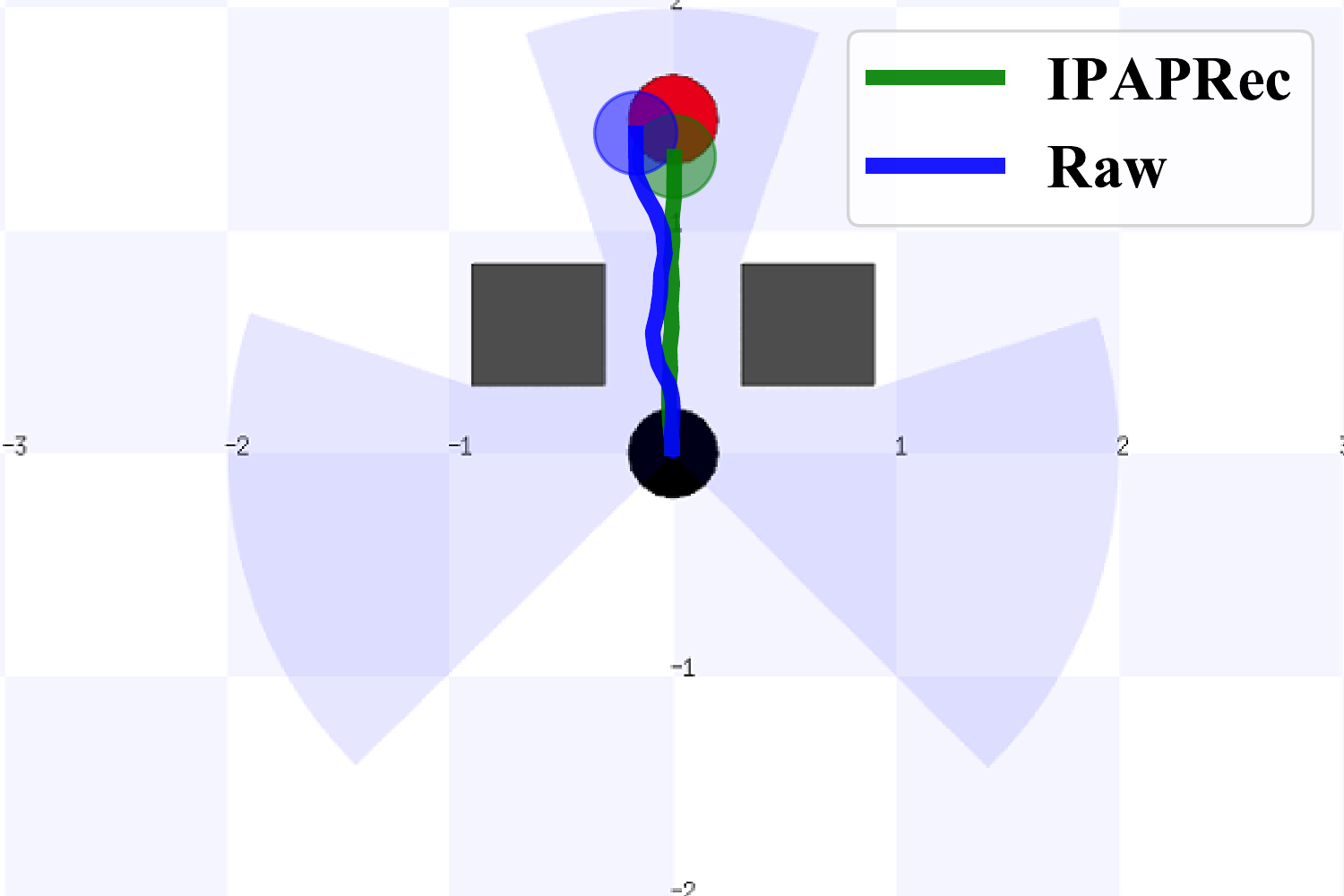}\label{evidence1}}
	  \subfloat[]{
        \includegraphics[width=0.40\linewidth]{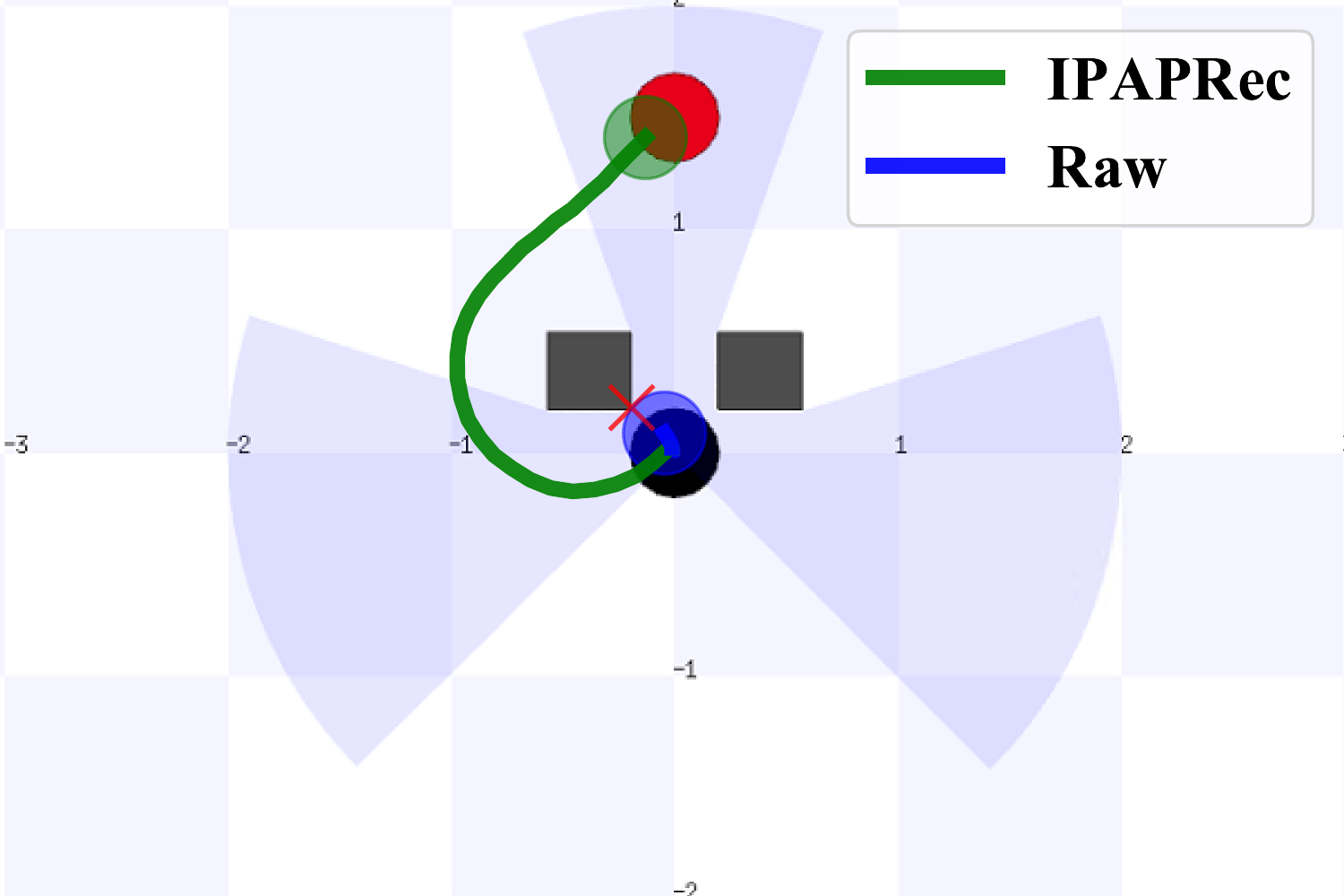}\label{evidence2}}
	\caption{Trajectories of agents trained with Raw and IPAPRec tested in scenarios introduced in Fig. \ref{limitation}.}
	\label{evidence}
\end{figure}

The training and testing details \hl{were} the same as in the first case, except that the replay buffer size, the total number of training steps, and model testing period \hl{were} increased to $2\times10^5$, $2\times10^5$, and $10^4$ steps, respectively. The learning curves are plotted in Fig. \ref{LC_4}. \hl{Since DWA and APF did not need training, their success rates in each test scenario remained unchanged with the increase of training steps.} As shown, both IPAPRec and IPAPRecN \hl{outperformed} their conventional counterparts in terms of success rate and learning speed in all scenarios. It is noteworthy that IPAPRec and IPAPRecN \hl{achieved} over 80\% success rate in Env\_3 due to the complex training scenario and new DNN model, which \hl{was} a great improvement compared to Case Study 1. \hl{Compared with APF and DWA, IPAPRec achieved much higher success rates in complex scenarios, such as Env\_3 and Env\_5.} To investigate the specific navigation behavior of the DRL agents in the open scenario, the trajectories of agents trained with Raw and IPAPRec tested in Env\_1 are plotted in Fig. \ref{traj}. As shown, in the empty scenario, the agent trained with IPAPRec \hl{navigated} with near-optimal paths (straight lines). In sharp contrast, the agent trained with Raw \hl{chose} much longer paths and sometimes even \hl{circled} around the targets, which \hl{was} consistent with the observations in \cite{Fan2020DistributedScenarios}. In addition, we \hl{tested} the agents trained with Raw and IPAPRec in the scenarios 
introduced in Fig. \ref{limitation}. As shown Fig. \ref{evidence1}, the agent trained with Raw \hl{moved} through the obstacle gate and reach its target, but it also \hl{tried} to move through the narrow obstacle gate in the other scenario (see Fig. \ref{evidence2}) and \hl{crashed} into the obstacle. In contrast, the agent trained with IPAPRec \hl{could} well distinguish the difference between the two scenarios and successfully \hl{completed} the two tasks.
\begin{table*}[t]
\renewcommand\arraystretch{1.2}
\caption{\label{tab:table-2} Maximum success rate (MSR) and average navigation score (MANS, as mean/SD) of the DRL agent in the training scenario and testing scenarios (Env1-3) trained with different IP methods.}
\centering
% \usepackage{multirow}
\setlength{\tabcolsep}{1.3mm}{\begin{tabular}{llllllllllll}
\hline
\hline
% \usepackage{multirow}
\multirow{3}{*}{Scenarios}                                                              & \multirow{3}{*}{IP Methods} & \multicolumn{10}{l}{Models (Training scenario)}                                                                                                                                                             \\
                                                                                        &                             & \multicolumn{2}{l}{Model\_0 (Env\_0)}  & \multicolumn{2}{l}{Model\_1 (Env\_4)} & \multicolumn{2}{l}{Model\_2 (Env\_4)} & \multicolumn{2}{l}{Model\_3 (Env\_5)}   & \multicolumn{2}{l}{Average}           \\
                                                                                        &                             & MSR (\%)          & MANS                & MSR (\%)          & MANS               & MSR (\%)           & MANS                 & MSR (\%)           & MANS                & MSR (\%)          & MANS               \\ \hline
\multirow{4}{*}{\begin{tabular}[c]{@{}l@{}}In   training \\      scenario\end{tabular}} & Raw                         & 59.6/6.4          & -0.36/0.07         & 97.2/1.0          & 0.23/0.04         & 56.8/13.7          & -0.51/0.16          & 85.2/7.9           & -0.05/0.16         & 74.7/19.1         & -0.17/0.31        \\
                                                                                        & LNorm                        & 34.4/11.1         & -0.67/0.11         & 96.0/1.8          & 0.25/0.03         & 88.4/6.5           & -0.06/0.15          & 38.0/16.0          & -0.69/0.15         & 64.2/30.0         & -0.29/0.42        \\
                                                                                        & IPAPRec                     & \textbf{86.4/2.7} & \textbf{0.02/0.02} & \textbf{98.0/0.0} & \textbf{0.36/0.0} & 92.8/4.7           & -0.01/0.12          & \textbf{100.0/0.0} & \textbf{0.29/0.01} & 94.3/5.9          & 0.17/0.17         \\
                                                                                        & IPAPRecN                    & 82.0/1.8          & -0.05/0.04         & \textbf{98.0/0.0} & \textbf{0.36/0.0} & \textbf{98.0/0.0}  & \textbf{0.26/0.01}  & \textbf{100.0/0.0} & 0.23/0.02 & \textbf{94.5/7.3} & \textbf{0.20/0.16} \\ \hline
\multirow{4}{*}{\begin{tabular}[c]{@{}l@{}}In   testing\\       scenarios\end{tabular}} & Raw                         & 58.8/2.2          & -0.35/0.04         & 71.3/4.6          & -0.13/0.06        & 42.5/6.5           & -0.59/0.11          & 76.3/2.8           & -0.09/0.07         & 62.2/13.7         & -0.29/0.22        \\
                                                                                        & LNorm                        & 36.5/2.7          & -0.64/0.05         & 70.3/3.0          & -0.13/0.05        & 65.9/13.9          & -0.3/0.17           & 37.6/7.1           & -0.63/0.08         & 52.6/17.5         & -0.42/0.24        \\
                                                                                        & IPAPRec                     & \textbf{88.7/1.7} & \textbf{0.06/0.02} & \textbf{92.1/0.9} & 0.19/0.01         & 75.1/3.8           & -0.23/0.09          & 92.8/1.3           & \textbf{0.22/0.01} & 87.2/7.5          & 0.06/0.19         \\
                                                                                        & IPAPRecN                    & 83.5/5.4          & -0.02/0.1          & 92.0/1.4          &\textbf{ 0.2/0.02 }& \textbf{91.6/2.0}  & \textbf{0.07/0.04}  & \textbf{94.3/0.9}  & 0.19/0.03 & \textbf{90.3/5.1} & \textbf{0.11/0.1}\\
\hline
\hline
\end{tabular}}
\end{table*}
\subsection{Overall performance comparison}
In this section, we \hl{summarized} the effect of IPAPRec and IPAPRecN on the DRL agents using the results obtained from the above three case studies. The DNN models to be evaluated are Model\_0, Model\_1, Model\_2 ($m=128$), and Model\_3. Two metrics named MSR (maximum success rate) and MANS (maximum average navigation score) \hl{were} utilized to evaluate the agent's navigation performance. MSR denotes the agent's maximum success rate during training. It measures the proportion of tasks that can be solved by the agent. MANS represents the maximum average navigation score received by the agent during training, and the score function \cite{zhang2021learn} is as follows, 
\begin{equation}
\begin{aligned}
S = 
\begin{cases}
1-\frac{2T_s}{T_{max}}, & \text{if success},\\
-1, & \text{otherwise},
\end{cases}
\end{aligned}
\end{equation}where $T_s$ denotes the navigation steps spent by the agent. This metric takes into account both navigation time and task completion, which provides a good measure of how fast an agent can complete a task. The comparison results based on these two metrics are summarized in Table \ref{tab:table-2}. As shown, the agents trained with IPAPRec and IPAPRecN substantially \hl{increased} the MSR and MANS values of their conventional counterparts in all cases. On average, in testing scenarios, IPAPRecN \hl{increased} the success rate of the agents trained with Raw from 62.2\% to 90.3\%, which is a remarkable improvement. It is noteworthy that IPAPRecN \hl{showed} a significant advantage when integrated with CNN models, while IPAPRec \hl{demonstrated} great performance improvement in terms of MANS for cases using FC models. 
\section{Real-world Performance Evaluation}\label{real-experiment}
In this section, we \hl{evaluated} the real-world performance of the DRL model trained with our IPAPRec method. The model used for testing \hl{was} Model\_3 (IPAPRec) introduced in Case Study 3 (see Section \ref{case3}), because it \hl{achieved} the highest MANS in simulation. For comparison, Model\_3 (Raw) \hl{was} selected as the benchmark. 

\subsection{Hardware setup}
\begin{figure}[t]
	\centering
	\includegraphics[width=0.4\linewidth]{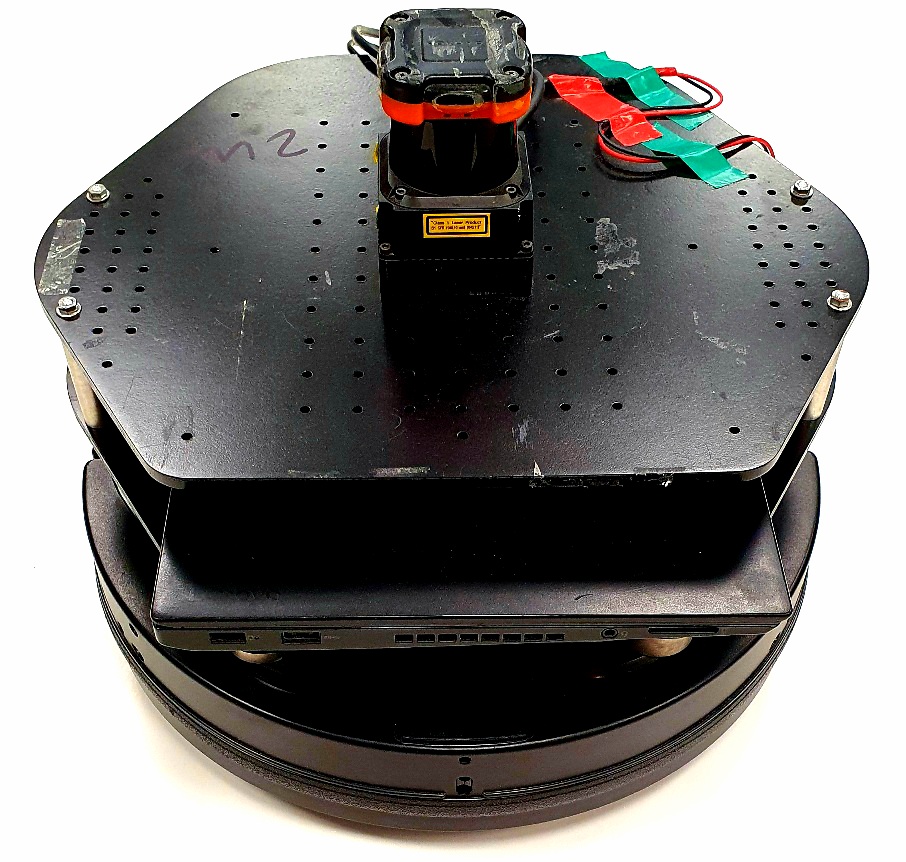}
	\caption{The Turtlebot2 robot used for testing.}
	\label{robot}
\end{figure}

As shown in Fig. \ref{robot}, the robot used for testing \hl{was} the Turtlebot2 robot. Its sensors and on-board computer are listed as follows,

\begin{itemize}
	\item \textit{LiDAR}: The adopted LiDAR hl{was} Hokuyo UTM-30LX LiDAR. It \hl{had} a FOV of $270^\circ$, a maximum measuring range of 30m, and an angular resolution of $0.25^\circ$.
	\item \textit{Target localization}: Since we \hl{did} not have a target localization sensor, such as the UWB localization system used in \cite{Fan2020DistributedScenarios}, the same as \cite{Tai2017,xie2021}, we \hl{pre-built} a map of the testing scenario for target localization. Specifically, we \hl{used} \textit{ROS GMapping} \cite{Gmapping} to build a map of the testing scenario and \hl{used} \textit{ROS AMCL} \cite{AMCL} to localize the robot in this map. The target position in robot frame \hl{could} be calculated using robot and goal coordinates in the map. It should be noted that this map \hl{was} only used to calculate the target position and \hl{was} not utilized for motion planning.
	\item \textit{On-board computer}: The on-board computer \hl{was} a laptop with an i7-7600U CPU. GPU \hl{was} not used.
\end{itemize}

\begin{figure}[t]
	\begin{minipage}{.4\linewidth}
	\subfloat[]{
       \centering\includegraphics[width=0.9\linewidth]{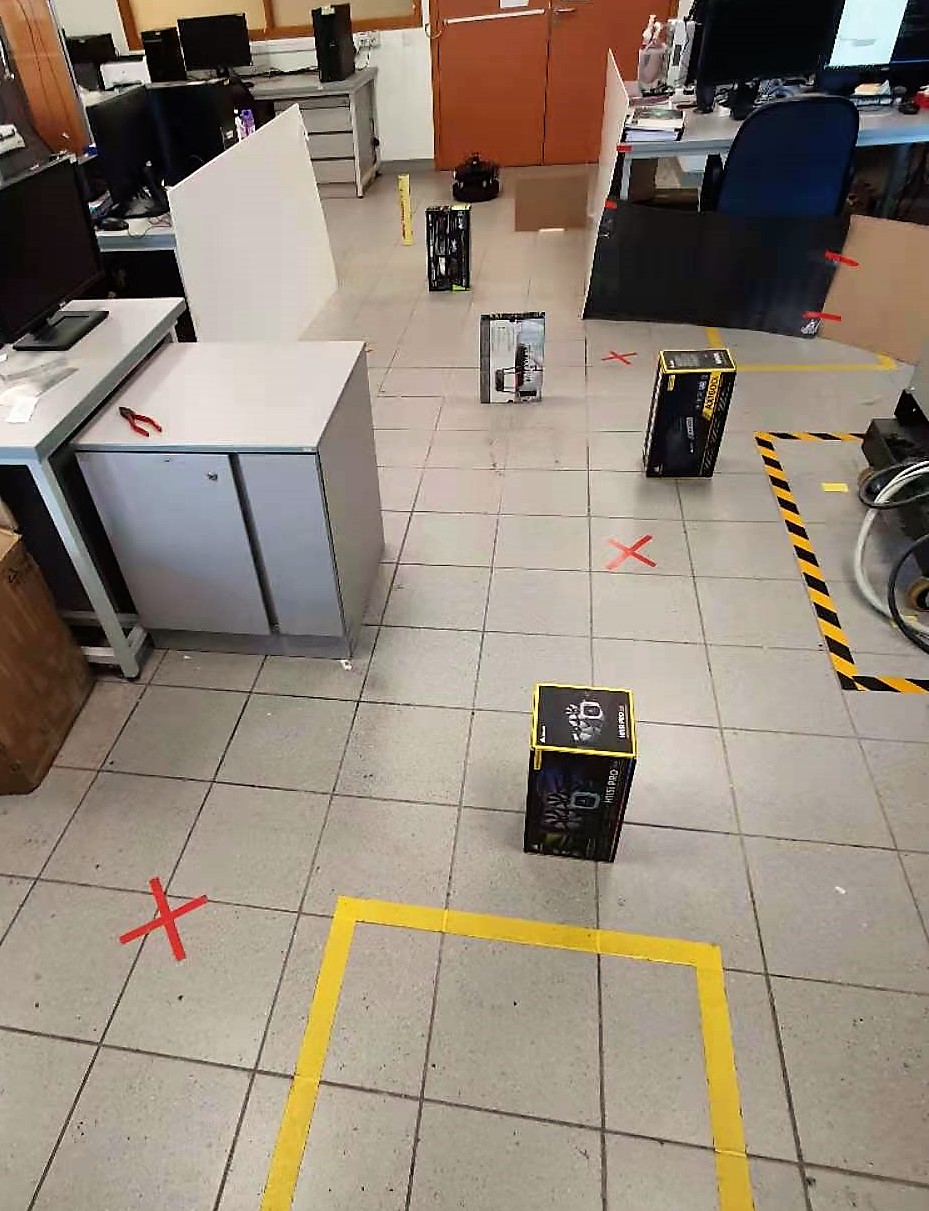}\label{REnv_0}}
	\end{minipage}%
	\begin{minipage}{.58\linewidth}
		\subfloat[]{
       \centering\includegraphics[width=0.9\linewidth]{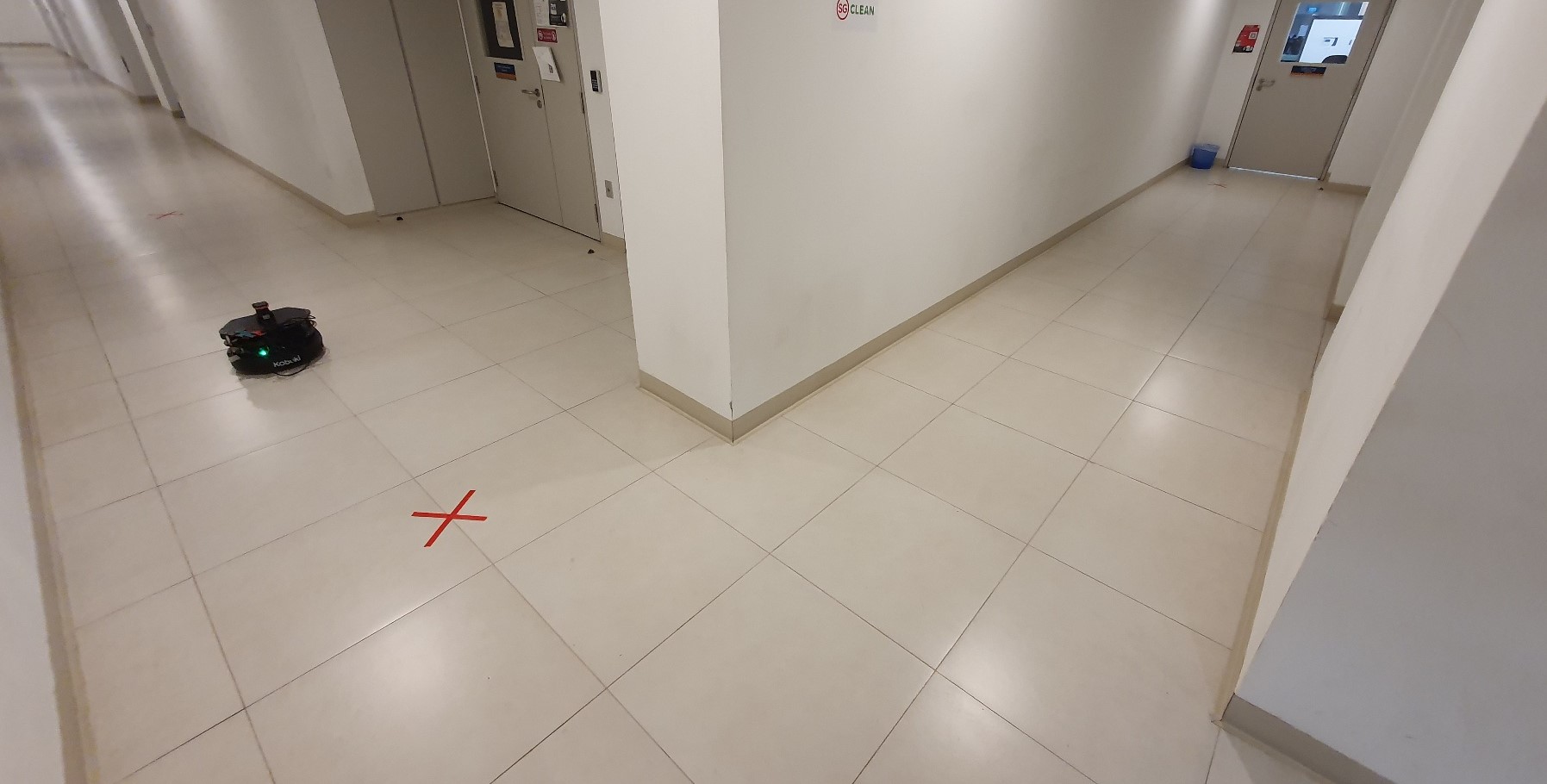}\label{REnv_1}}\\[-2ex]
       	\subfloat[]{
       \centering\includegraphics[width=0.9\linewidth]{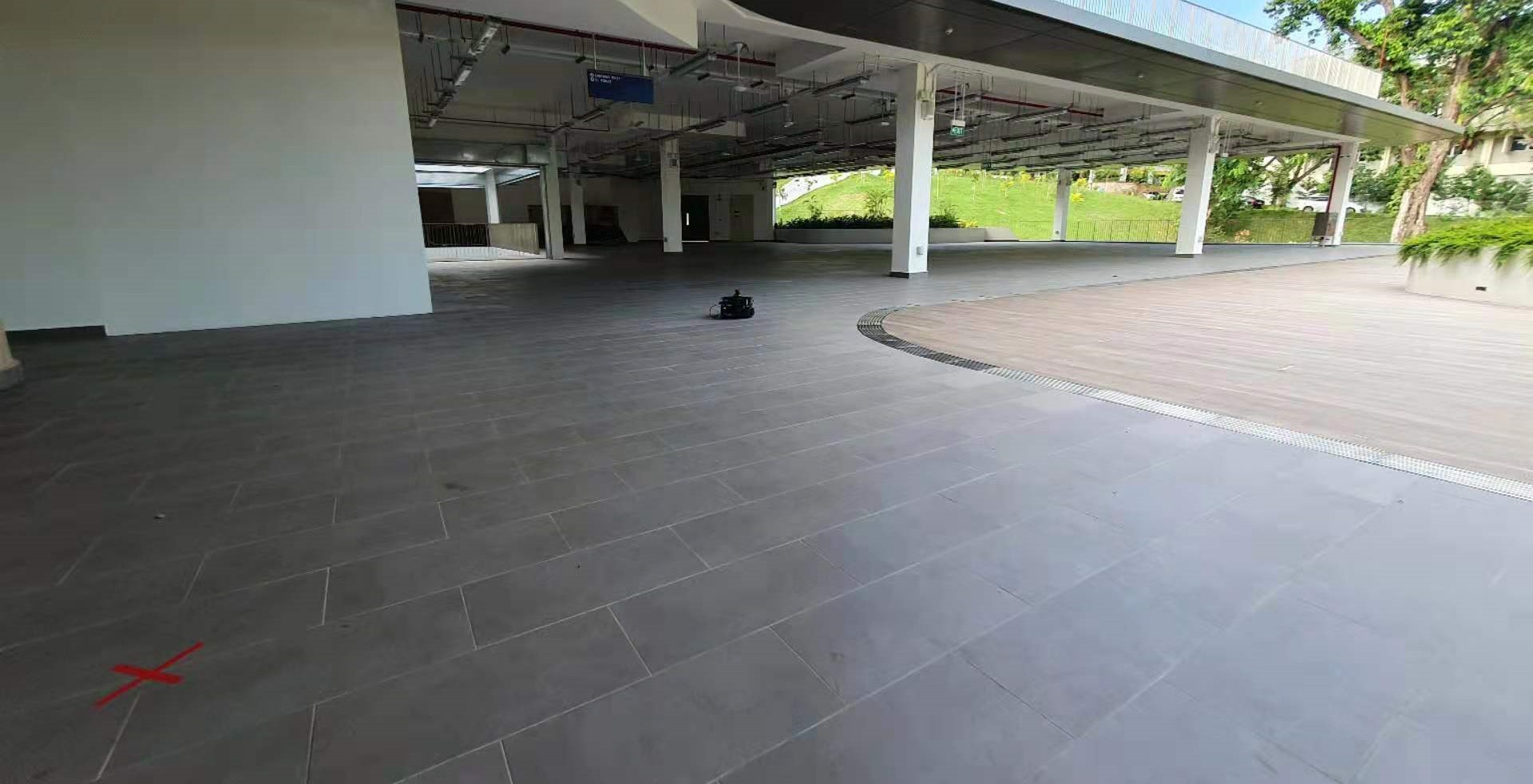}\label{REnv_2}}
	\end{minipage}
	\caption{Real-world testing scenarios. (a) REnv\_0: a crowded indoor  scenario; (b) REnv\_1: a corridor scenario; (c) REnv\_2: an outdoor scenario.}
	\label{REnv}
\end{figure}
\subsection{Testing scenarios and task description}
The robot \hl{was} tested in three real-world scenarios shown in Fig. \ref{REnv}. The three scenarios, \hl{including} a crowded indoor scenario, a corridor scenario \hl{and} an outdoor scenario, \hl{differed} widely in the degree of \hl{crowdedness} and \hl{were} good choices to evaluate the generalization performance of the DRL agents. In each testing scenario, as shown in Fig. \ref{real_test}, the robot \hl{started} from the point  ``S" and \hl{needed} to reach several targets in succession. The targets \hl{were} labelled as  $``\text{G}i"$ and marked with red crosses markers on the floor (see Fig. \ref{REnv}). \hl{After reaching one target, the robot would wait for two seconds to show it successfully reached this target  rather than occasionally traversing this target when moving to the other target.} If the robot \hl{failed} to reach its target in the former task, it \hl{would} be placed at the former target point and \hl{faced} the new target before the new task \hl{started}. 
\subsection{Experimental Results}
\begin{figure}[t!]
	\centering\subfloat[Raw in REnv\_0]{
	\centering\includegraphics[width=0.45\linewidth]{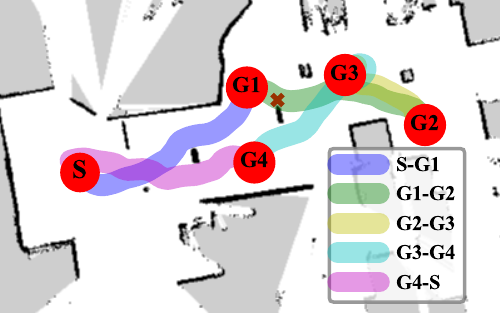}\label{t1}}
	\hfill\subfloat[IPAPRec in REnv\_0]{
	\centering\includegraphics[width=0.45\linewidth]{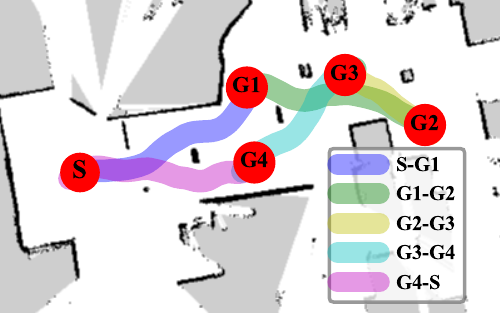}\label{t2}}
	\\[-2ex]
	\subfloat[Raw in REnv\_1]{
	\centering\includegraphics[width=0.45\linewidth]{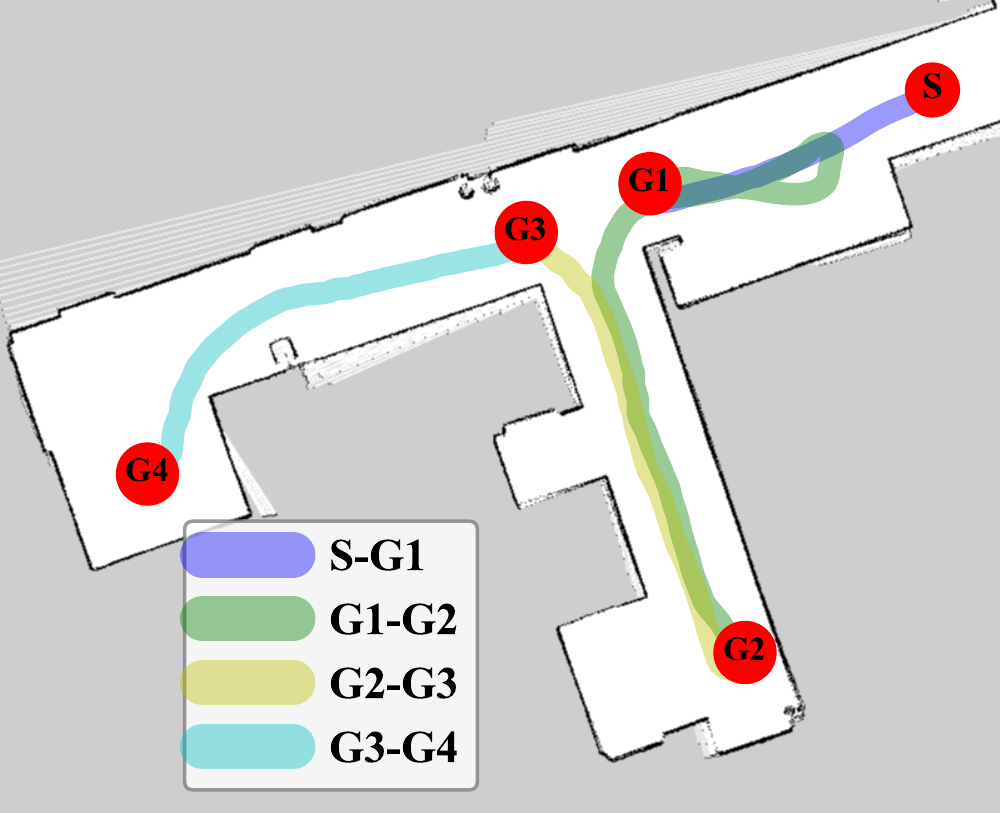}\label{t3}}
	\hfill\subfloat[IPAPRec in REnv\_1]{
	\centering\includegraphics[width=0.45\linewidth]{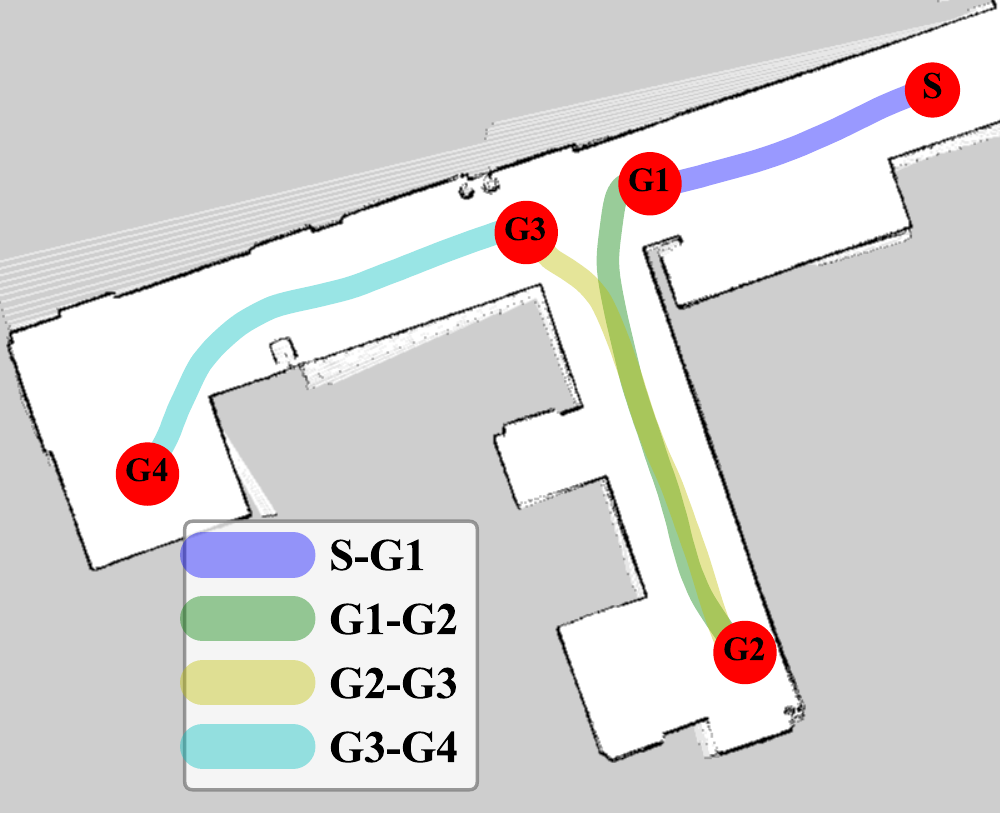}\label{t4}}
	\\[-2ex]
	\subfloat[Raw in REnv\_2]{
	\centering\includegraphics[width=0.45\linewidth]{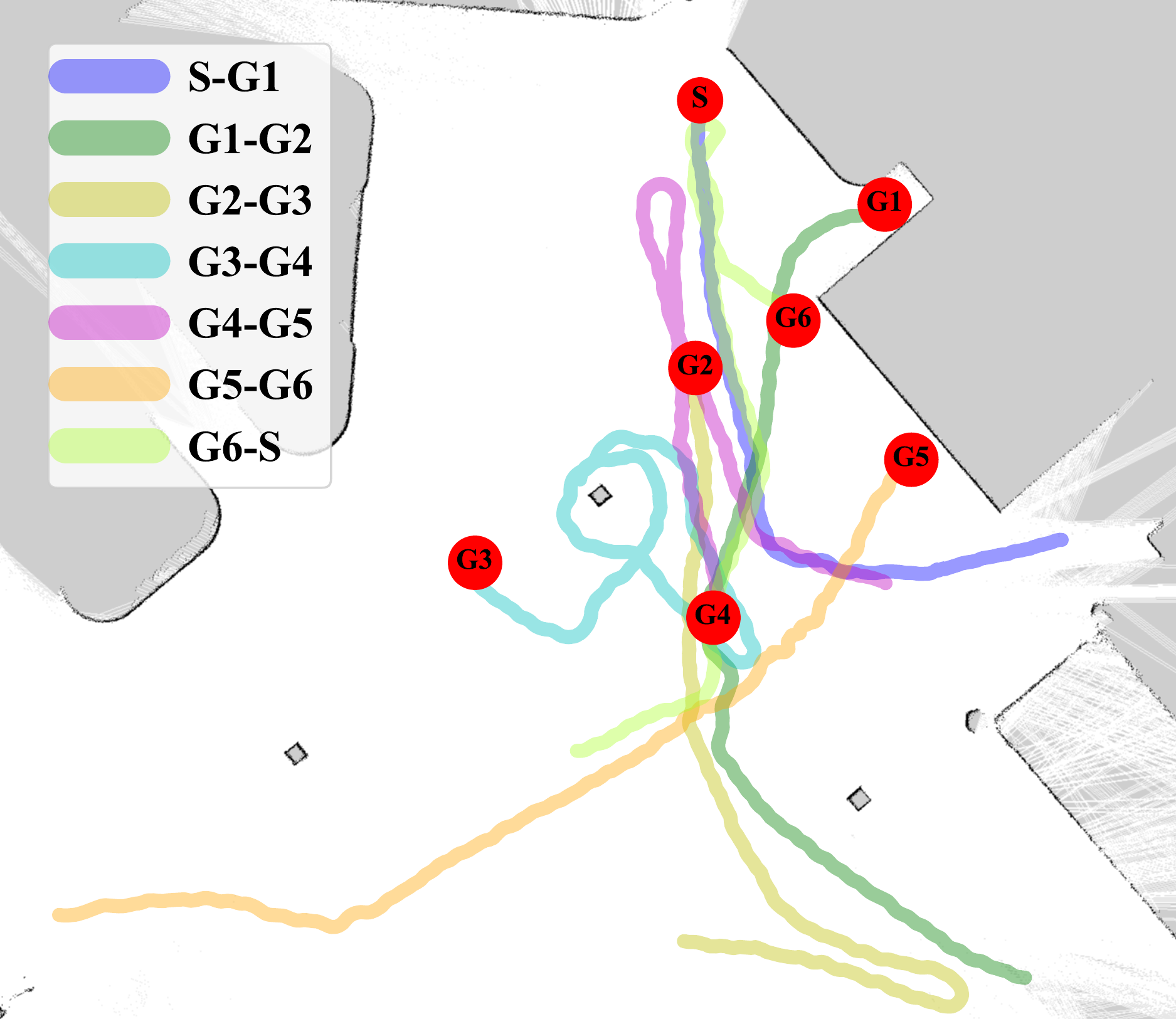}\label{t5}}
	\hfill\subfloat[IPAPRec in REnv\_2]{
	\centering\includegraphics[width=0.45\linewidth]{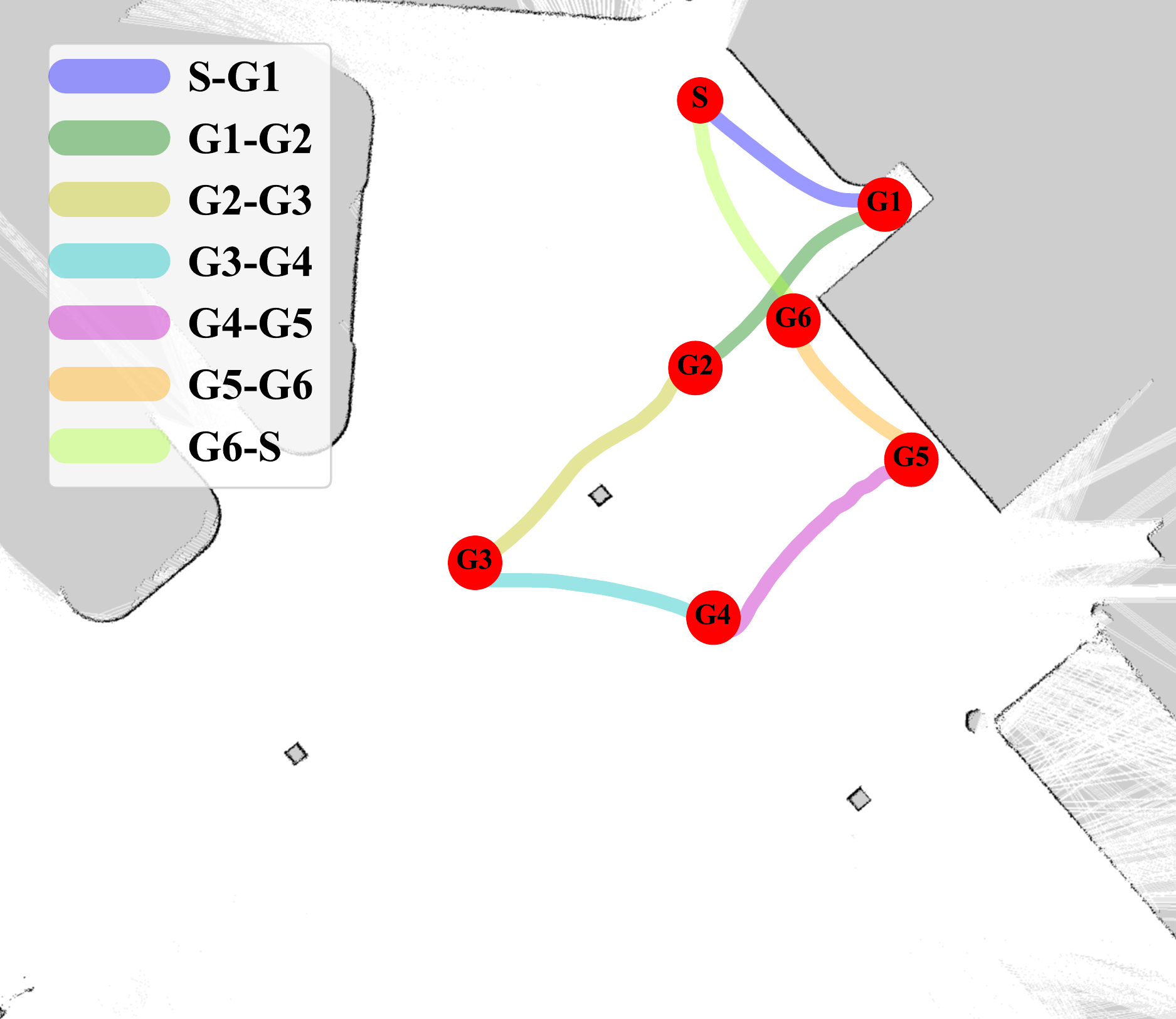}\label{t6}}
	\caption{Trajectories of robot trained with Raw and IPAPRec when tested in indoor, corridor and outdoor scenarios. The  experimental videos can be found in the supplementary file \hl{or in {\href{https://youtu.be/slRbEXioGd4}{https://youtu.be/slRbEXioGd4}.}}}
	\label{real_test}
\end{figure}

The robot trajectories and testing results are plotted in Fig. \ref{real_test} and summarized in Table \ref{tab:table-3}, respectively. As shown, trained with IPAPRec, the DRL agent \hl{completed} all the tasks in the three scenarios and the resulted trajectories \hl{were} near-optimal and smooth. In shape contrast, the agent trained with Raw \hl{crashed} in the crowded scenario REnv\_0 ($\text{G}1\rightarrow \text{G}2$), \hl{generated} jagged trajectories in REnv\_1 and \hl{failed} to reach its destinations in REnv\_2. Moreover, as shown in Table  \ref{tab:table-3}, compared to Model\_3 (Raw), the total navigation time and distance \hl{were} considerably reduced when IPAPRec is applied. Furthermore, we \hl{assessed} the agility of the robot controlled by Model\_3 (IPAPRec). As shown in the videos, dynamic obstacles \hl{would} suddenly appear in front of the robot during its navigation. and the robot \hl{could} quickly respond to unforeseen changes and successfully \hl{completed} the tasks.

\begin{table}[t!]
\renewcommand\arraystretch{1.2}
\caption{\label{tab:table-3} Testing results in three real-world scenarios. The IPAPRec method is compared with Raw in terms of total navigation distance (TND), total navigation time (TNT) and number of completed tasks (NCT).}
\centering
% \usepackage{multirow}
\setlength{\tabcolsep}{1.4mm}{\begin{tabular}{lllllll}
\hline
\hline
\multirow{2}{*}{Metric} & \multicolumn{2}{l}{In REnv\_0} & \multicolumn{2}{l}{In REnv\_1} & \multicolumn{2}{l}{In REnv\_2} \\
                        & Raw       & IPAPRec            & Raw          & IPAPRec         & Raw       & IPAPRec            \\ \hline
TND (m)                 & 13.8      & \textbf{13.6}      & 35.9         & \textbf{29.7}   & 179.9     & \textbf{45.3}      \\
TNT (s)                 & 51.8      & \textbf{37.9}      & 105.9        & \textbf{68.0}   & 420.6     & \textbf{110.7}     \\
NCT                     & 4         & \textbf{5}         & \textbf{4}   & \textbf{4}      & 0         & \textbf{7}      \\   
\hline
\hline
\end{tabular}}
\end{table}

\section{Conclusion}\label{conclusion}
In this work, we discuss why conventional input representation of LiDAR readings can cause performance degradation of DRL agents and propose IPAPRec/IPAPRecN approaches to address this issue. With the LiDAR representations processed by IPAPRec/IPAPRecN, the DRL agent can learn navigation skills efficiently. Extensive experiments have been carried out, and the comparison results verify that the  IPAPRec/IPAPRecN approaches can significantly increase the learning speed of the DRL agent and substantially enhance the agent's performance in unseen crowded or open scenarios. In testing scenarios, IPAPRecN improves the highest success rate from 62.2\% to 90.3\%. \hl{Due to the excellent performance and ease of implementation, in the future, we plan to apply our IPAPRec method to robots navigating in 3D scenarios, such as quadrotors.}

% or


\section*{Acknowledgment}
Wei Zhang would like to thank the financial support from China Scholarship Council and National University of Singapore.

\ifCLASSOPTIONcaptionsoff
  \newpage
\fi



%\bibliographystyle{IEEEtran}

\bibliographystyle{IEEEtran}
\bibliography{references}



%\vfill

\end{document}